\documentclass{article}
\PassOptionsToPackage{numbers}{natbib}
 \usepackage{graphicx}
\usepackage{booktabs}
\usepackage{multirow}
\usepackage{graphicx}
\usepackage{amsmath}

\usepackage[preprint]{neurips_2025}

\usepackage[utf8]{inputenc} 
\usepackage[T1]{fontenc}    
\usepackage{hyperref}       
\usepackage{url}            
\usepackage{booktabs}       
\usepackage{amsfonts}       
\usepackage{nicefrac}       
\usepackage{microtype}      
\usepackage{xcolor}         
\usepackage[letterpaper]{geometry}
\usepackage{subcaption}
\usepackage[table]{xcolor}
\usepackage{graphicx}
\usepackage{changes}

\title{Rhamba: Region-Aware Hybrid Attention–Mamba Framework for Self-Supervised Learning in Resting-State fMRI}

\author{%
Pankaj Pandey$^{1,*}$ \quad 
Ruthwik Reddy Doodipala$^{1,2,*}$ \quad 
Pratheek Eranki$^{1,2}$ \quad \\
\textbf{Carolina Torres-Rojas}$^{1}$ \quad
\textbf{Manob Jyoti Saikia}$^{2}$ \quad 
\textbf{Ranganatha Sitaram}$^{1}$ \\ \\
$^{1}$St. Jude Children's Research Hospital \quad 
$^{2}$The University of Memphis\\
\texttt{\{rdoodipa,ppandey,rsitaram\}@stjude.org} \\
$^{*}$Co-first authors; names are presented in chronological order.
}

\begin{document}
\maketitle
\begingroup
\addtocounter{footnote}{-1}
\endgroup

\begin{abstract}

Self-supervised pretraining is promising for large-scale neuroimaging representation learning, yet the impact of region-aware masking and hybrid sequence modeling remains underexplored. In this work, we introduce \textbf{Rhamba}, a region-aware pretraining framework that integrates anatomically guided masking with hybrid Attention–Mamba architectures for resting state functional magnetic resonance imaging (fMRI) analysis. Models were pretrained on the ABIDE dataset using region-aligned patch embeddings and three masking strategies with increasing spatial specificity. We evaluated four architectural variants: a Mamba-dominant model, an Alternate architecture with interleaved Mamba and Attention blocks, and two hybrid encoder–decoder configurations (Attention–Mamba (AM) and Mamba–Attention (MA)). The pretrained models were fine-tuned on downstream classification tasks using the COBRE and ADHD-200 datasets for schizophrenia and attention-deficit/hyperactivity disorder discrimination.  We employed Integrated Gradients, an explainable AI method, to identify the brain regions contributing to model predictions. Our masking strategy strongly influenced reconstruction behavior, with reconstruction loss following a consistent ordering (\textit{Any} $>$ \textit{Majority} $>$ \textit{Pure}). However, this trend did not directly translate into downstream performance, where differences were modest and dataset-dependent. The hybrid architecture with the MA configuration achieved the highest average AUROC across both datasets. Rhamba demonstrated competitive performance against state-of-the-art methods. Region-wise analysis showed that peak performance depends on the interaction between masking strategy and architecture rather than a single dominant configuration. The hybrid architectures underscore the importance of combining global context modeling with efficient sequence dynamics. Overall, Rhamba offers a flexible framework for balancing interpretability, scalability, and performance in large-scale fMRI representation learning.
\end{abstract}

\clearpage
\section{Introduction}
Resting-state functional magnetic resonance imaging (rs-fMRI) provides a non-invasive measure of intrinsic large-scale brain organization by capturing spontaneous fluctuations in the blood-oxygen-level-dependent (BOLD) signal, an indirect marker of neural activity reflecting changes in regional cerebral blood flow and oxygenation \cite{ogawa1990brain,belliveau1991functional}. In the absence of explicit tasks, correlated low-frequency BOLD dynamics reveal functional connectivity patterns that characterize distributed brain networks \cite{biswal1995functional,damoiseaux2006consistent,biswal2012resting}. Over the past decade, large-scale data-sharing initiatives such as the Autism Brain Imaging Data Exchange (ABIDE I and II) \cite{di2014autism} and ADHD-200 \cite{adhd2012adhd} have substantially expanded the availability of multi-site rs-fMRI datasets spanning diverse populations and neurodevelopmental conditions. Despite this increased data availability, most rs-fMRI studies remain dataset-specific, relying on handcrafted features such as seed-based correlations or region-averaged connectivity matrices, conventional statistical modeling, or supervised approaches trained on limited labeled samples \cite{zhang2025common,hilbert2024lack,biswal2025history,arbabshirani2017single,varoquaux2018cross}.

Recent progress in deep learning has further advanced rs-fMRI analysis by enabling end-to-end modeling of high-dimensional brain signals. A wide range of supervised architectures, including convolutional neural networks (CNNs) \cite{meszlenyi2017resting,qureshi20193d,gulhan2025use}, recurrent neural networks (RNNs) and long short-term memory (LSTM) models \cite{dvornek2017identifying,guo2022characterization}, graph neural networks (GNNs) \cite{wang2025novel,zhang2022classification}, encoder–decoder frameworks \cite{kim2021representation}, and more recently Transformer-based models \cite{dai2024classification,kwon2025predicting} have demonstrated strong performance across tasks such as disease classification and behavioral prediction. These models are capable of capturing complex spatiotemporal dependencies in rs-fMRI data and often outperform traditional machine learning approaches when sufficient labeled data are available. However, the effectiveness of supervised deep learning in neuroimaging remains fundamentally constrained by the scarcity and heterogeneity of labeled datasets. Clinical annotations are expensive and time-consuming to obtain, cohort sizes are often modest, and variability across sites, scanners, and preprocessing pipelines further limits generalization \cite{varoquaux2018cross,specht2020current}. As a result, many supervised models exhibit reduced robustness when evaluated across independent datasets or real-world settings, hindering their scalability and broader applicability \cite{huf2014generalizability,hilbert2024lack,marek2022reproducible}.

These limitations have motivated a shift toward self-supervised learning, where models are pretrained on large collections of unlabeled rs-fMRI data using proxy objectives and subsequently adapted to downstream tasks \cite{kim2023swiftswin4dfmri,dong2024brainjepabraindynamicsfoundation}. Self-supervised learning (SSL) for resting-state fMRI has evolved along several distinct paradigms, each leveraging different inductive biases to learn meaningful brain representations without labeled data. Early approaches largely adopted contrastive learning on functional connectivity graphs, where models such as unsupervised contrastive graph learning (UCGL) and its extensions learned representations by maximizing agreement between augmented views of brain networks \cite{wang2023unsupervised}. These methods rely on data augmentation strategies applied to BOLD signals or graph structures, which, while effective, may inadvertently alter biologically meaningful temporal dependencies or connectivity patterns \cite{you2020graph,zhang2023gcl,wang2025self}. To address this, more recent work introduces diffusion-based augmentation to better preserve signal integrity while perturbing connectivity patterns \cite{wang2025self}. In parallel, reconstruction-based approaches, particularly masked autoencoders, learn to recover missing portions of 4D fMRI data, enabling joint modeling of spatial and temporal dependencies through transformer architectures \cite{gao20253d}. An alternative direction explores proxy-task supervision, where predicting subject identity enforces subject-discriminative representations, although such strategies may bias models toward individual-specific rather than disease-relevant features \cite{hashimoto2021deep}. Complementary efforts optimize intrinsic neurobiological criteria, such as functional homogeneity, to derive personalized brain networks, improving interpretability while potentially constraining representational flexibility \cite{li2023computing}. Additionally, mutual information-based frameworks capture spatiotemporal dynamics by maximizing dependencies between local and global temporal contexts \cite{mahmood2020whole}. These approaches highlight the diversity of self-supervised objectives for fMRI representation learning.

Beyond the design of self-supervised objectives, recent work has explored a diverse set of architectural paradigms for modeling fMRI data. Early approaches primarily relied on CNNs to learn representations directly from voxel-level inputs, effectively capturing local spatial structure. CNN-based encoder–decoder models learn personalized functional networks through self-supervised objectives grounded in functional homogeneity \cite{li2023computing}, while convolutional–recurrent architectures combined with contrastive learning capture spatiotemporal patterns from raw fMRI data \cite{jaiswal2023detecting}. Building on this, graph-based methods represent the brain as a functional network, enabling graph neural networks to model inter-regional dependencies. Self-supervised graph learning frameworks, including masked graph autoencoding and graph convolutional network (GCN) based approaches, improve representation learning and disease prediction by jointly modeling network topology and temporal dynamics \cite{wen2023graph,wang2024graph}. Transformer-based architectures further advance this direction by enabling the capture of global interactions and long-range dependencies across both spatial and temporal dimensions. SwiFT extends Swin transformer \cite{liu2021swintransformerhierarchicalvision} architecture to 4D fMRI through windowed self-attention, enabling scalable spatiotemporal representation learning \cite{kim2023swiftswin4dfmri}, while transformer-based foundation models trained with masked autoencoding demonstrate strong generalization across downstream tasks \cite{ferrante2025self}. However, the quadratic complexity of self-attention limits their ability to efficiently process long fMRI sequences. To address these scalability limitations, state-space models (SSMs), particularly Mamba, enable linear-time sequence modeling and efficient handling of long temporal contexts. In fMRI, Mamba-based approaches improve the modeling of global spatiotemporal dependencies while reducing computational overhead \cite{deng2025causal, mamba2026state, Wang2026NeuroSTORM}, as demonstrated by NeuroSTORM, which leverages a Mamba backbone for large-scale foundation modeling. Nevertheless, purely state-space models lack the flexibility of attention mechanisms for capturing complex global interactions. Consequently, recent work increasingly adopts hybrid architectures that integrate transformers and state-space models, combining efficient temporal modeling with expressive global attention. Hybrid Transformer-Mamba designs achieve improved scalability and performance in sequence modeling \cite{lieber2024jambahybridtransformermambalanguage,li2026transmambasequencelevelhybridtransformermamba,park2024mambalearnlearncomparative,waleffe2024empiricalstudymambabasedlanguage}, and similar approaches in neuroimaging jointly capture long-range temporal dynamics and spatial dependencies in fMRI data \cite{zhao2025research,kannan2025brainmt}.

Despite these advances in both learning objectives and architectural design, existing self-supervised approaches largely lack explicit integration of neuroanatomical structure. The brain exhibits a spatially structured and functionally specialized organization, where distinct regions contribute differently to cognition and disease processes. In response, a growing body of work has begun incorporating region-level priors into representation-learning frameworks. Recent approaches introduce region-aware modeling through Region of Interest (ROI) embeddings and graph attention mechanisms to capture inter-regional relationships and functional specificity \cite{yang2024brainmae}. These methods are motivated by the observation that brain regions exhibit consistent functional roles and structured connectivity patterns across individuals, enabling more interpretable and biologically grounded representations. Similarly, region-guided masking strategies have been explored in reconstruction-based pretraining, where anatomically defined regions are selectively masked to encourage learning of localized and functionally meaningful signals \cite{doodipala2025region}. Such approaches demonstrate that masking coherent anatomical regions, rather than random voxels, improves both representation quality and downstream performance. However, existing region-aware methods remain limited in several key aspects. Many approaches operate on ROI-level representations or predefined graph structures, potentially sacrificing fine-grained spatial information, while others incorporate anatomical priors without fully leveraging flexible spatiotemporal sequence modeling. Moreover, current masking strategies lack systematic control over the granularity and consistency of anatomical perturbations.

\quad To address these limitations, we propose \textbf{Rhamba}, a \emph{Region-Aware Hybrid Attention–Mamba Architecture} for large-scale resting-state fMRI modeling. Rhamba is built upon three complementary design principles. First, it incorporates \emph{region-aware modeling}, where anatomically defined brain regions guide tokenization and masking during self-supervised pretraining of resting-state 4D volumes. By structuring the learning process around neuroanatomical organization, this design provides a principled foundation for interpretability while enabling the model to capture structured inter-regional dependencies beyond generic voxel-level representations. Second, Rhamba employs a \emph{hybrid backbone} that interleaves state-space (Mamba) layers with self-attention blocks, combining efficient long-range temporal modeling with explicit global contextual integration across distributed brain networks. This hybrid design addresses the limitations of purely attention-based or state-space approaches by jointly modeling extended temporal dynamics and global functional interactions. Third, Rhamba is trained using volumetric self-supervised objectives directly on full-resolution resting-state fMRI data, enabling scalable and transferable representations across heterogeneous cohorts and acquisition sites.

By integrating anatomical priors with hybrid sequence modeling, Rhamba establishes a unified framework that bridges interpretability and scalability in large-scale resting-state neuroimaging.

Our contributions are summarized as follows:
\begin{itemize}
    \item We introduce \textbf{Rhamba}, a region-aware hybrid Attention-Mamba architecture tailored for large-scale resting-state fMRI model.
    \item We propose anatomically guided tokenization and masking strategies that leverage structural priors from MNI-aligned Automated Anatomical Labeling (AAL) atlas parcellations, enabling region-aware patch embeddings and masking during volumetric self-supervised pretraining of resting-state 4D brain volumes.
    \item We demonstrate that interleaving state-space and attention mechanisms enables efficient long-sequence modeling while preserving global functional interactions in resting-state fMRI data.
    \item We benchmark the proposed framework against established fMRI foundation models, including SwiFT \cite{kim2023swiftswin4dfmri} and NeuroSTORM \cite{Wang2026NeuroSTORM}, and observe competitive performance across downstream classification tasks for schizophrenia and attention-deficit/hyperactivity disorder discrimination.
\end{itemize}

\section{Methods}
\subsection{Self-supervised pretraining Framework}
We design a two-stage learning pipeline consisting of self-supervised pretraining followed by supervised fine-tuning. In the pretraining phase, the model is optimized to reconstruct missing portions of the input fMRI sequence, while the fine-tuning stage adapts the learned representations for downstream prediction tasks.

During pretraining, we employ a region-aware masking strategy applied to patch-based representations of the 4D fMRI volumes. Each patch is aligned with anatomically defined regions of interest (ROIs), allowing masking decisions to be guided by neuroanatomical structure. Instead of masking tokens uniformly at random across space and time, we selectively remove spatiotemporal patches based on their regional assignment. This encourages the model to recover structured brain activity patterns within specific regions, promoting the learning of spatially localized and functionally meaningful features.

The model is trained using a reconstruction objective that predicts the masked portions of the input sequence without requiring labeled data. This setup enables the encoder to capture rich spatiotemporal dependencies present in resting-state fMRI signals. For downstream tasks, the decoder used during pretraining is discarded, and a lightweight prediction head is attached to the encoder. The resulting model is then fine-tuned end-to-end using task-specific supervision. This separation between representation learning and task adaptation allows the framework to leverage large-scale unlabeled data while maintaining flexibility for different applications.

\begin{figure}[!htbp]
  \centering
  \includegraphics[width=1\linewidth]{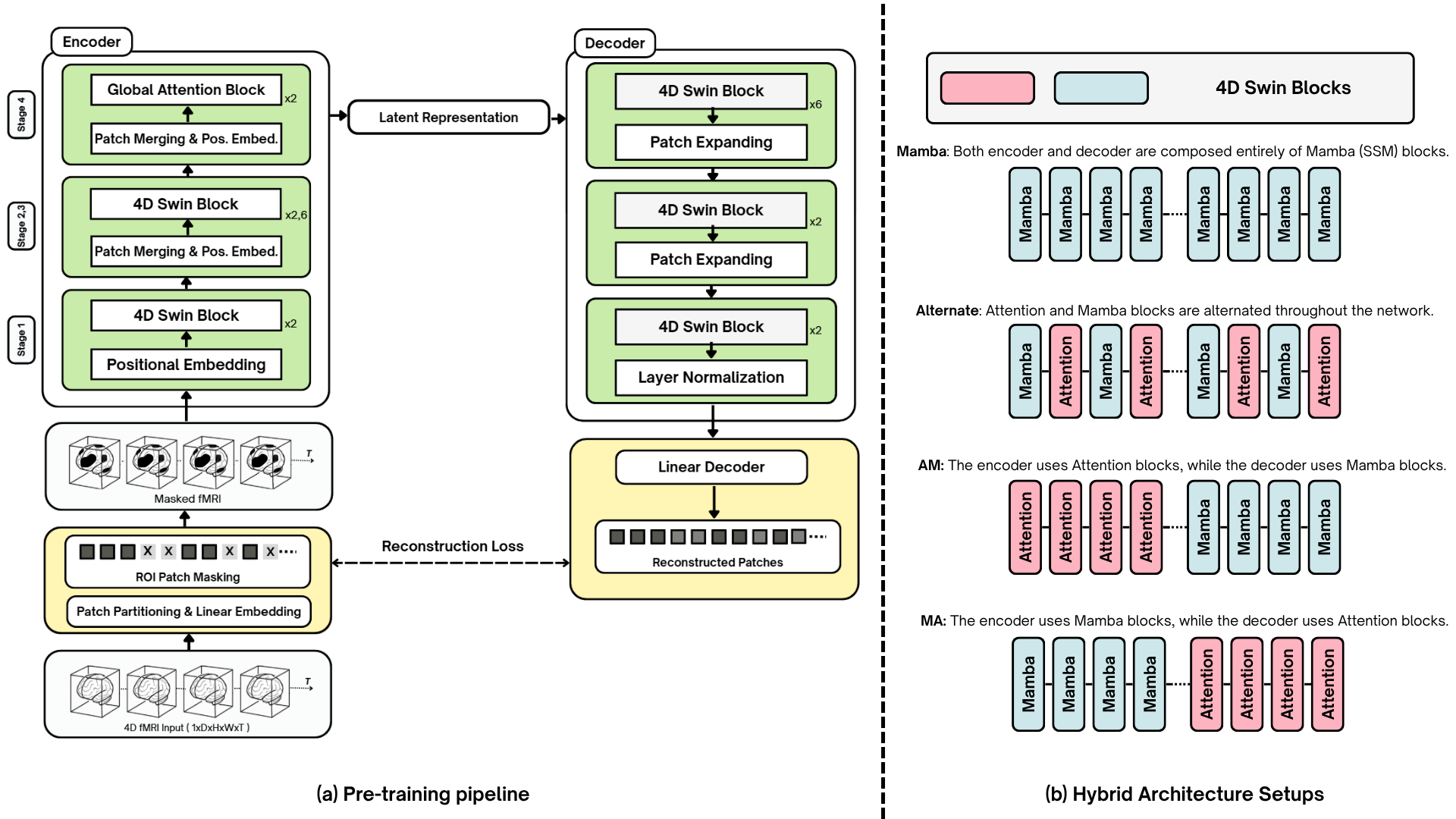}
  \caption{Overview of the proposed framework. (a) Pre-training pipeline, including ROI-based  patch embedding, patch masking, a hierarchical 4D Swin encoder–decoder architecture, and the reconstruction objective. (b) Hybrid block configurations, where each block corresponds to a 4D Swin Transformer block using either Attention or Mamba (SSM), including Mamba, Alternate, Attention-Mamba (AM), and Mamba-Attention (MA) setups. In the implemented encoder, the final hierarchical stage includes full self-attention before the latent representation.}
  \label{fig:roi-masking}
\end{figure}

As the backbone, we utilize NeuroSTORM \cite{Wang2026NeuroSTORM}, a foundation model designed for direct modeling of full 4D fMRI volumes. NeuroSTORM is pretrained on a large-scale collection of over 28 million fMRI frames from more than 50,000 subjects spanning datasets such as UK Biobank \cite{UKB}, ABCD \cite{ABCD}, and HCP \cite{HCP}. Its architecture is based on a Shifted-Window Mamba (SWM) design, which integrates windowed spatial partitioning with state-space sequence modeling to efficiently capture both temporal dynamics and spatial structure.
 Specifically, NeuroSTORM integrates Mamba \cite{mamba}, a linear-time state-space model, to efficiently capture long-range temporal dependencies, with Swin Transformer blocks \cite{liu2021swintransformerhierarchicalvision} that employ shifted-window self-attention to model spatial relationships \cite{kim2023swiftswin4dfmri}.

Swin blocks operate by partitioning the input 3D volume into non-overlapping local windows and applying self-attention within each window, significantly reducing the computational cost compared to global attention. To enable interaction across neighboring regions, the window configuration is shifted between successive layers, allowing information to propagate beyond local windows while maintaining efficiency. This hierarchical and locality-aware design is particularly well-suited for high-dimensional fMRI data, where both fine-grained local structure and broader spatial context are important.

As shown in (Figure.~\ref{fig:roi-masking}), the encoder consists of multiple hierarchical stages that progressively reduce spatial resolution through patch merging while increasing feature dimensionality. Each stage is composed of stacked 4D Swin-style blocks, where the core operation within each block is instantiated using either self-attention or Mamba (state-space) layers. The final encoder stage includes a full self-attention block before the latent representation. This design preserves the efficiency of Mamba-based sequence modeling in earlier stages while allowing the deepest representation to integrate global contextual information across the reduced token space. Thus, the encoder combines efficient long-range temporal modeling with explicit global interaction at the final stage, where attention is computationally more tractable after spatial downsampling. The decoder mirrors the encoder and progressively restores spatial resolution through patch expansion layers. It is similarly composed of stacked 4D Swin-style blocks with either attention or Mamba operators, enabling effective integration of contextual information during reconstruction. A final linear projection maps the decoded features back to voxel space to produce the reconstructed 4D fMRI volume.

\subsubsection{Hybrid Attention--Mamba Architecture Experimentation}
\label{sec:hybrid}
State-space models (SSMs) such as Mamba provide computationally efficient modeling 
of long spatiotemporal sequences and are particularly well suited for 
high-dimensional fMRI data. However, large-scale empirical analyses have 
shown that pure SSM-based architectures may underperform attention-based 
models in tasks requiring precise contextual integration, information 
routing, and in-context learning \cite{waleffe2024empiricalstudymambabasedlanguage}. 
Hybrid Transformer–Mamba designs have therefore been proposed to combine 
the global interaction modeling capacity of attention with the efficiency 
of SSMs \cite{waleffe2024empiricalstudymambabasedlanguage, 
li2026transmambasequencelevelhybridtransformermamba}. 
In medical imaging applications, hybrid Mamba architectures have also been 
reported to improve performance when modeling complex spatial dependencies 
\cite{bansal2025comprehensivesurveymambaarchitectures}.

Given the distributed and hierarchical nature of brain activity patterns, 
we investigate whether integrating attention mechanisms into the 
NeuroSTORM masked autoencoding framework improves fMRI representation 
learning. To this end, we design four encoder–decoder variants that differ in the 
relative placement and proportion of attention and Mamba blocks across 
the encoder and decoder (Figure.~\ref{fig:roi-masking}). These configurations include a Mamba-dominant backbone, an alternating architecture in which attention and Mamba blocks are interleaved across the network, and asymmetric encoder--decoder variants in which one component is predominantly attention-based while the other is predominantly Mamba-based. In all encoder variants, the final hierarchical stage includes a full self-attention block before the latent representation. This final attention stage is applied after progressive spatial downsampling, where the token sequence is shorter and global contextual integration is computationally more tractable. Thus, the AM and MA variants should be interpreted as attention- or Mamba-dominated configurations rather than strictly exclusive encoder--decoder decompositions.

All configurations maintain identical encoder and decoder depth, hidden 
dimension, patching strategy, masking protocol, and optimization schedule 
to ensure a controlled comparison.  Each hybrid configuration is evaluated using masked voxel reconstruction 
error (voxel-level mean squared error on masked regions) during self-supervised 
pretraining, followed by downstream task performance using a lightweight prediction head. This controlled ablation allows us to assess whether global attention is more beneficial in the encoder or decoder, whether sparse insertion of attention layers suffices, and how these design choices affect downstream generalization for fMRI-based prediction tasks.

\subsection{Region-aware fMRI Pretraining Dataset and Preprocessing}
\label{abide_pretraining}
Resting-state fMRI data from the Autism Brain Imaging Data Exchange (ABIDE) \cite{di2014autism} consortium were used for large-scale self-supervised pretraining. Raw 4D NIfTI volumes were processed to ensure anatomical consistency, harmonized spatial and temporal resolution, and quantitative quality control across sites. For each subject, spatial dimensions, voxel resolution, repetition time (TR), and affine transformations were extracted from image headers. To assess anatomical alignment in native space, a subject-specific mean fMRI volume was computed across time. A standard MNI152 brain mask was affinely resampled to the native fMRI grid, and a subject-specific brain mask was estimated from the functional data. Anatomical correspondence was quantified using the Dice similarity coefficient between the subject mask and the projected MNI mask, providing a robust measure of global spatial alignment independent of voxel grid differences.

All fMRI volumes were spatially normalized to the canonical MNI152 template at 2\,mm isotropic resolution using affine-aware interpolation. An Automated Anatomical Labeling (AAL3) atlas was resampled to the same MNI grid using nearest-neighbor interpolation to preserve discrete regional labels. To standardize temporal resolution across acquisition sites, time series were resampled to a fixed TR of 0.8\,s along the temporal dimension while preserving spatial geometry. To obtain a uniform representation suitable for deep learning, volumes were cropped to a fixed spatial field-of-view of $96 \times 96 \times 96$ voxels. The same spatial transformation was applied to the resampled AAL atlas, and affine matrices were updated to maintain accurate millimeter-space correspondence. The resulting representation consisted of MNI-aligned 2\,mm isotropic volumes of size $96^3 \times T$.

Intensity normalization was performed using global z-scoring within the brain mask. Specifically, mean and standard deviation were computed across all brain voxels and timepoints, and voxel intensities were standardized according to
\begin{equation}
x' = \frac{x - \mu}{\sigma}.
\end{equation}
To reduce the influence of extreme values, normalized intensities were clipped to the range $[-5, 5]$, and voxels outside the brain mask were set to zero. Across subjects, the 1st–99th percentile intensity range after normalization typically fell within approximately $[-4, 1.6]$, indicating stable distributional properties across sites.

\subsubsection{Inclusion and Exclusion Criteria}

Subjects were retained for pretraining if they satisfied two quality criteria. First, anatomical alignment was required to satisfy
\begin{equation}
\mathrm{Dice}(\text{subject mask}, \text{MNI mask}) > 0.85.
\end{equation}
Second, distributional outliers were identified using the 99th percentile of normalized brain intensities ($p_{99}$). The distribution of subject-level ($p_{99}$) values across subjects was used to define an interquartile range–based threshold, and subjects were excluded if their $p_{99}$ exceeded this threshold, defined as
\begin{equation}
p_{99} \leq Q_3 + 1.5 \times IQR,
\end{equation}
where $Q_1 = 1.2646$, $Q_3 = 1.5132$, and $IQR = 0.2487$, yielding a threshold of 1.8862.

Based on these criteria, 44 subjects were excluded due to intensity outliers, 16 due to insufficient anatomical alignment, and 1 for failing both conditions. The final pretraining cohort comprised 1{,}041 subjects.

\begin{table}
\centering
\caption{\textbf{Macroregion-level patch sets under three anatomical criteria.}
For a $16 \times 16 \times 16$ grid (4096 patches) with 216 voxels per patch,
we report the number of patches selected and the corresponding voxel footprint
if all selected patches are masked (patches $\times 216$).}
\label{tab:macroregion_patch_sets}
\small
\setlength{\tabcolsep}{6pt}
\begin{tabular}{lrrrrrr}
\hline
\textbf{Macroregion} &
\multicolumn{2}{c}{\textbf{Any}} &
\multicolumn{2}{c}{\textbf{Majority}} &
\multicolumn{2}{c}{\textbf{Pure}} \\
\cline{2-3}\cline{4-5}\cline{6-7}
& \textbf{Patches} & \textbf{Masked voxels} &
  \textbf{Patches} & \textbf{Masked voxels} &
  \textbf{Patches} & \textbf{Masked voxels} \\
\hline
Frontal      & 524 & 113{,}184 & 242 & 52{,}272 & 140 & 30{,}240 \\
Parietal     & 319 &  68{,}904 & 135 & 29{,}160 &  93 & 20{,}088 \\
Occipital    & 233 &  50{,}328 &  96 & 20{,}736 &  47 & 10{,}152 \\
Temporal     & 334 &  72{,}144 & 133 & 28{,}728 &  82 & 17{,}712 \\
Limbic       & 259 &  55{,}944 &  56 & 12{,}096 &  23 &  4{,}968 \\
Subcortical  & 104 &  22{,}464 &  26 &  5{,}616 &  12 &  2{,}592 \\
Cerebellum   & 233 &  50{,}328 & 115 & 24{,}840 &  54 & 11{,}664 \\
\hline
\end{tabular}
\end{table}

\subsubsection{Patch-Level Anatomical Classification}

For spatial modeling, each $96^3$ volume was partitioned into non-overlapping cubic patches of size $6 \times 6 \times 6$ voxels (216 voxels per patch), yielding a $16 \times 16 \times 16$ grid (4096 patches). Using the aligned AAL3 atlas, patch-level anatomical coverage was quantified for seven macro-anatomical systems (frontal, parietal, occipital, temporal, limbic, subcortical, and cerebellum). Detailed anatomical compositions for each system are provided in Supplementary Table 1. Patch membership was defined using three progressively stricter criteria:

\paragraph{Any-labeled patches.}
A patch was assigned to a macro-region if it contained at least one voxel whose AAL label belonged to that region:
\[
\exists \, v \in P \;\text{such that}\; v \in \mathcal{R}.
\]
This definition captures spatial overlap irrespective of boundary mixing.

\paragraph{Majority-labeled patches.}
A stricter criterion required that more than 50\% of the 216 voxels within a patch belonged to the macro-region:
\[
\frac{|P \cap \mathcal{R}|}{216} > 0.5.
\]
This ensures that the patch is predominantly occupied by the region of interest.

\paragraph{Pure patches.}
For majority-labeled patches, the dominant AAL label was identified. A patch was classified as \emph{pure} if the dominant label accounted for at least 70\% of labeled voxels within the patch:
\[
\frac{\max_k n_k}{n_{\mathrm{labeled}}} \geq 0.70,
\]
where $n_k$ denotes the voxel count of label $k$ and $n_{\mathrm{labeled}}$ is the number of non-zero voxels in the patch. The purity criterion isolates spatially coherent, anatomically homogeneous patches.

\begin{figure}[t]
  \centering
  \includegraphics[width=1\linewidth]{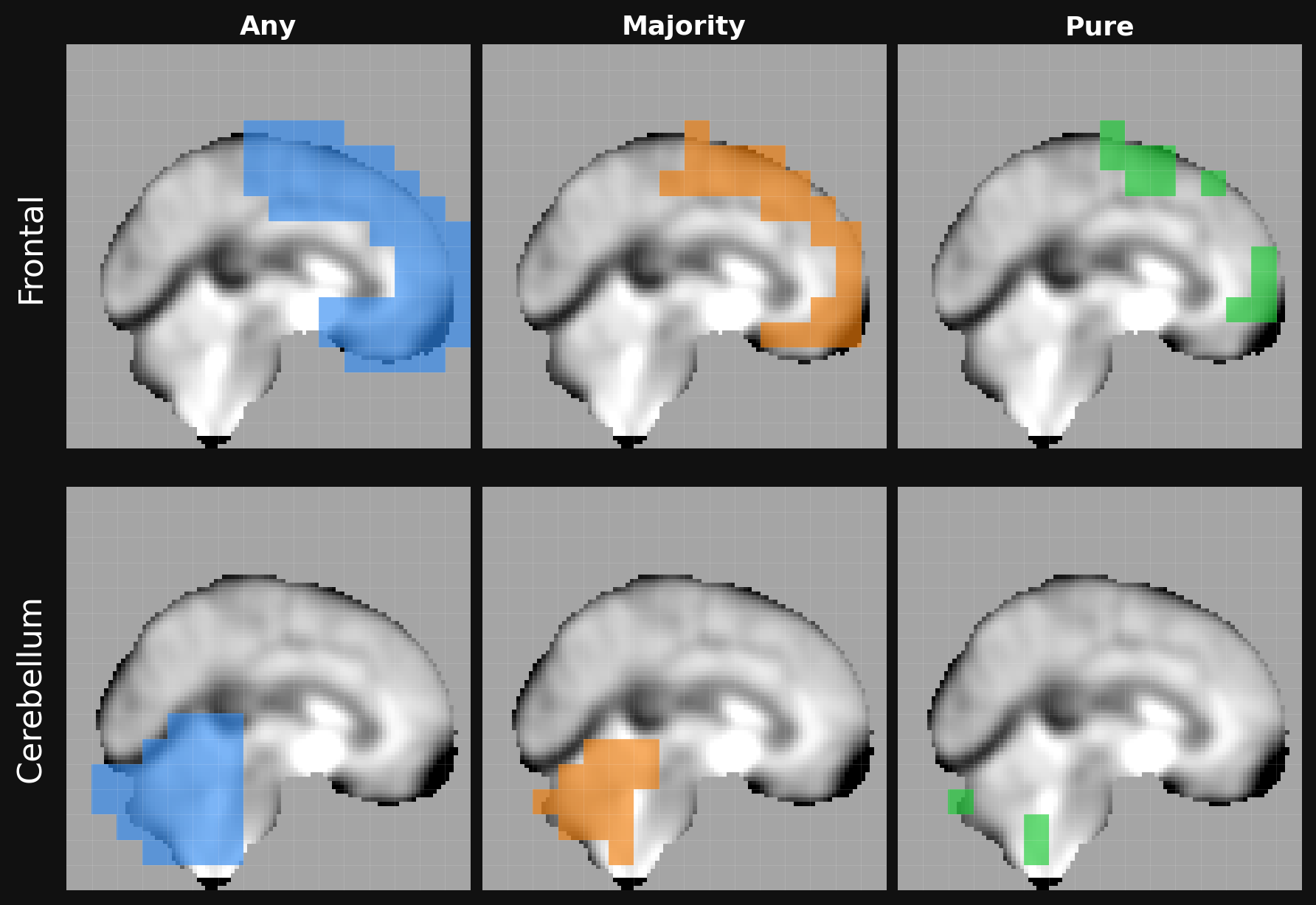}
  \caption{ Representative sagittal-view masking visualizations. Visualization of the three anatomical masking configurations applied to the frontal lobe (top row) and cerebellum (bottom row) during masked autoencoder (MAE) pretraining. The brain volume ($96^3$) is divided into a $16 \times 16 \times 16$ grid of $6 \times 6 \times 6$ voxel patches. For each patch, inclusion is determined by its overlap with the AAL-atlas region: \textit{Any}  includes patches with at least one labeled region voxel; \textit{Majority} includes patches where more than $ 50\%$ of voxels belong to the target region; and \textit{Pure} includes majority-labeled patches where the dominant AAL label accounts for at least $70\%$ of the labeled (non-zero) voxels within the patch. For the frontal lobe, this yields 524, 242, and 140 patches respectively; for the cerebellum, 233, 115, and 54 patches respectively. Sagittal slices through each region's centroid are shown on a representative ABIDE subject.}
  \label{fig:roi-masking_representation}
\end{figure}

Across macro-regions, the three criteria provide complementary perspectives: the \emph{any} criterion measures anatomical coverage, the \emph{majority} criterion measures dominant spatial occupancy, and the \emph{pure} criterion isolates homogeneous anatomical units. Patch counts and corresponding voxel footprints under each definition are summarized in Table~\ref{tab:macroregion_patch_sets}. All masking strategies used during pretraining operated on this fixed 4096-patch grid, ensuring consistent region-level control across subjects. Figure \ref{fig:roi-masking_representation} illustrates the three region-assignment configurations for the frontal lobe and cerebellum.

\subsection{Downstream Datasets and Evaluation Protocol}

We utilize two publicly available resting-state fMRI datasets, COBRE and ADHD-200, in the main manuscript, with MND included as an additional validation dataset in the Supplementary Materials to evaluate downstream performance across both controlled and heterogeneous acquisition settings. All datasets were processed using the same pipeline described in Section~\ref{abide_pretraining}, including spatial normalization to MNI152 space (2\,mm isotropic), temporal resampling to a fixed TR of 0.8\,s, global intensity normalization, and patch-wise representation on a fixed $96^3$ grid with AAL3-based anatomical alignment.

\paragraph{COBRE dataset.}
The Center for Biomedical Research Excellence (COBRE)  dataset \cite{Bellec2015} consists of 146 subjects, including 72 individuals with schizophrenia and 74 healthy controls, with an age range of 18 - 65 years. The dataset is relatively homogeneous in acquisition and is widely used for benchmarking schizophrenia classification. Consistent with its controlled acquisition setting, anatomical alignment with the MNI template was high and stable across subjects (Dice: $0.94 \pm 0.01$).

\paragraph{ADHD-200 dataset.}
The ADHD-200 dataset \cite{adhd200} consists of 973 subjects, including 362 individuals with ADHD and 611 healthy controls, with an age range of of 7 - 21 years. The dataset is collected across multiple sites with varying acquisition protocols, introducing significant heterogeneity that makes it well-suited for assessing how well models generalize across different scanners and population groups. In contrast to COBRE, the ADHD-200 cohort exhibits substantially greater variability in anatomical alignment (Dice: $0.62 \pm 0.12$), reflecting its multi-site heterogeneity.

\paragraph{Quality control strategy.}
In contrast to pretraining, no exclusion based on Dice similarity or intensity-based outlier detection was applied to the downstream datasets. While such criteria improve anatomical and distributional consistency, they may disproportionately remove valid samples in heterogeneous multi-site cohorts such as ADHD-200. Retaining the full dataset allows evaluation under realistic conditions, where variability in acquisition, subject motion, and scanner characteristics is intrinsic. This design enables a more rigorous assessment of model robustness and generalizability across diverse clinical settings. Across both datasets, the downstream task was binary classification (case vs control), using representations derived from the pretrained model under a consistent preprocessing and patchification framework.
\subsection{Interpretability Method}
To interpret model predictions, we employed Integrated Gradients with SmoothGrad Squared (IG-SQ), implemented using the Captum framework \cite{sundararajan2017axiomaticattributiondeepnetworks}. Attribution maps were computed for samples in the held-out test set, and only correctly classified subjects were included in the analysis to ensure reliability of the explanations.

For each subject, IG attributions were squared to obtain IG-SQ maps, which emphasize high-magnitude contributions while reducing the effect of noise. The resulting 4D attribution maps were then spatially smoothed using a Gaussian filter applied independently at each time point. Subsequently, the maps were averaged along the temporal dimension to obtain a single 3D attribution map per subject. To improve robustness, this procedure was repeated across five independent training runs with different random seeds. All subject-level maps were aligned to a common anatomical space and aggregated across seeds and subjects. Prior to aggregation, maps were globally normalized to account for scale differences across runs. A group-level attribution map was then obtained by averaging the normalized maps, capturing consistent explanatory patterns. 

For visualization, the group-level map was thresholded by retaining only the top percentile of attribution values, highlighting the most salient brain regions associated with the prediction task.
\subsection{Experimental Settings}
Self-supervised pretraining was conducted on the ABIDE dataset, comprising 1,041 subjects. The model was trained for 30 epochs with a batch size of 8 and approximately 7.7 million trainable parameters. We used the AdamW optimizer \cite{kingma2014adam} with a learning rate of $5 \times 10^{-5}$. The training objective was the mean squared error (MSE) between the reconstructed outputs and the masked regions of the input fMRI volumes. All experiments were run on a high-performance computing cluster equipped with four/six NVIDIA H100 GPUs, each with 80\,GB of HBM3 memory, enabling efficient large-scale training.

For downstream evaluation, the dataset was divided at the subject level into training, validation, and test sets in an 8:1:1 ratio. We compared Rhamba with representative state-of-the-art architectures from both the transformer-based paradigm, SwiFT \cite{kim2023swiftswin4dfmri}, and the state-space paradigm, NeuroSTORM \cite{Wang2026NeuroSTORM}. For NeuroSTORM, we evaluated three masking strategies that combine different spatial and temporal masking schemes: \textit{Random + Random}, \textit{Window + Random}, and \textit{Random + Tube}. The first term refers to the spatial masking strategy: \textit{Random} masks patches uniformly across the volume, whereas \textit{Window} masks contiguous patches. The second term refers to the temporal masking strategy: \textit{Random} masks time points independently, whereas \textit{Tube} masks the same spatial locations across all time points. To ensure a fair comparison, all experiments used identical subject-level train--validation--test partitions, the same preprocessing pipeline, and the same pretraining dataset. Downstream fine-tuning was performed under matched optimization and evaluation settings where applicable. For each architecture, region, and masking configuration, checkpoint selection was based only on validation performance; the selected checkpoint was then evaluated once on the held-out test set. The main comparative results in Section \ref{comparison_results} are reported over five independent runs.

\subsection{Model-free analysis of masking-configuration difficulty}
To explain why reconstruction loss differs across the three anatomically defined masking configurations (\textit{Any}, \textit{Majority}, and \textit{Pure}), we characterized the intrinsic BOLD-signal structure of the masked patches independently of the pretrained model. Each $96 \times 96 \times 96$ volume was partitioned into a $16 \times 16 \times 16$ grid of 4096 non-overlapping patches, with each patch containing $6 \times 6 \times 6$ voxels, matching the tokenization used during pretraining. For each patch, we obtained a scalar time series by spatially averaging its voxels and then z-scored each time series across time. The analysis used resting-state fMRI data from 500 ABIDE subjects, with 100 time points per subject.

For each of the seven anatomical regions, the \textit{Any}, \textit{Majority}, and \textit{Pure} configurations selected progressively more restrictive sets of masked patches. To remove masked-set size as a potential confound, we equalized the number of masked patches across configurations within each region using a matched budget corresponding to the minimum patch count across the three configurations: Frontal, 140; Parietal, 93; Temporal, 82; Cerebellum, 54; Occipital, 47; Limbic, 23; and Subcortical, 12. Larger patch sets were randomly subsampled to the corresponding budget over five independent draws, and the resulting values were averaged.

For each region $\times$ configuration combination, we computed four candidate explanatory metrics: (i) \textit{within-masked correlation}, defined as the mean pairwise Pearson correlation among the masked-patch time series and interpreted as the intrinsic redundancy of the target set; (ii) \textit{masked--visible correlation}, defined as the mean correlation between each masked patch and its six nearest visible brain patches and interpreted as proximity to the surrounding context; (iii) \textit{mean pairwise distance}, defined as the average Euclidean distance between masked-patch centers in grid units and interpreted as spatial spread; and (iv) \textit{signal variance}, defined as the mean temporal variance of the masked patches and interpreted as signal amplitude. Each metric was averaged across subjects, and 95\% confidence intervals were estimated using bootstrap resampling of subjects with 2000 resamples. We related each model-free metric to reconstruction loss across the 21 region × configuration conditions using Spearman rank correlation.

\subsection{Statistical Analysis}

Statistical analyses were performed to assess differences between masking strategies for both downstream performance (AUROC) and reconstruction loss. Since the same set of region–architecture configurations was evaluated under each masking strategy, comparisons were treated as repeated measures. For global comparisons across the three masking strategies (\textit{Any}, \textit{Majority}, and \textit{Pure}), we used the non-parametric Friedman test, which is appropriate for related samples without assuming normality. Post hoc pairwise comparisons were conducted using the Wilcoxon signed-rank test to evaluate differences between individual masking strategies. To account for multiple comparisons, Bonferroni correction was applied. All statistical tests were two-sided. Corrected $p$-values less than $0.05$ were considered statistically significant. Additionally, $p$-values less than $0.06$ were reported as trend-level significance to highlight marginal effects.
\begin{figure}[t]
  \centering
  \includegraphics[width=1.\linewidth]{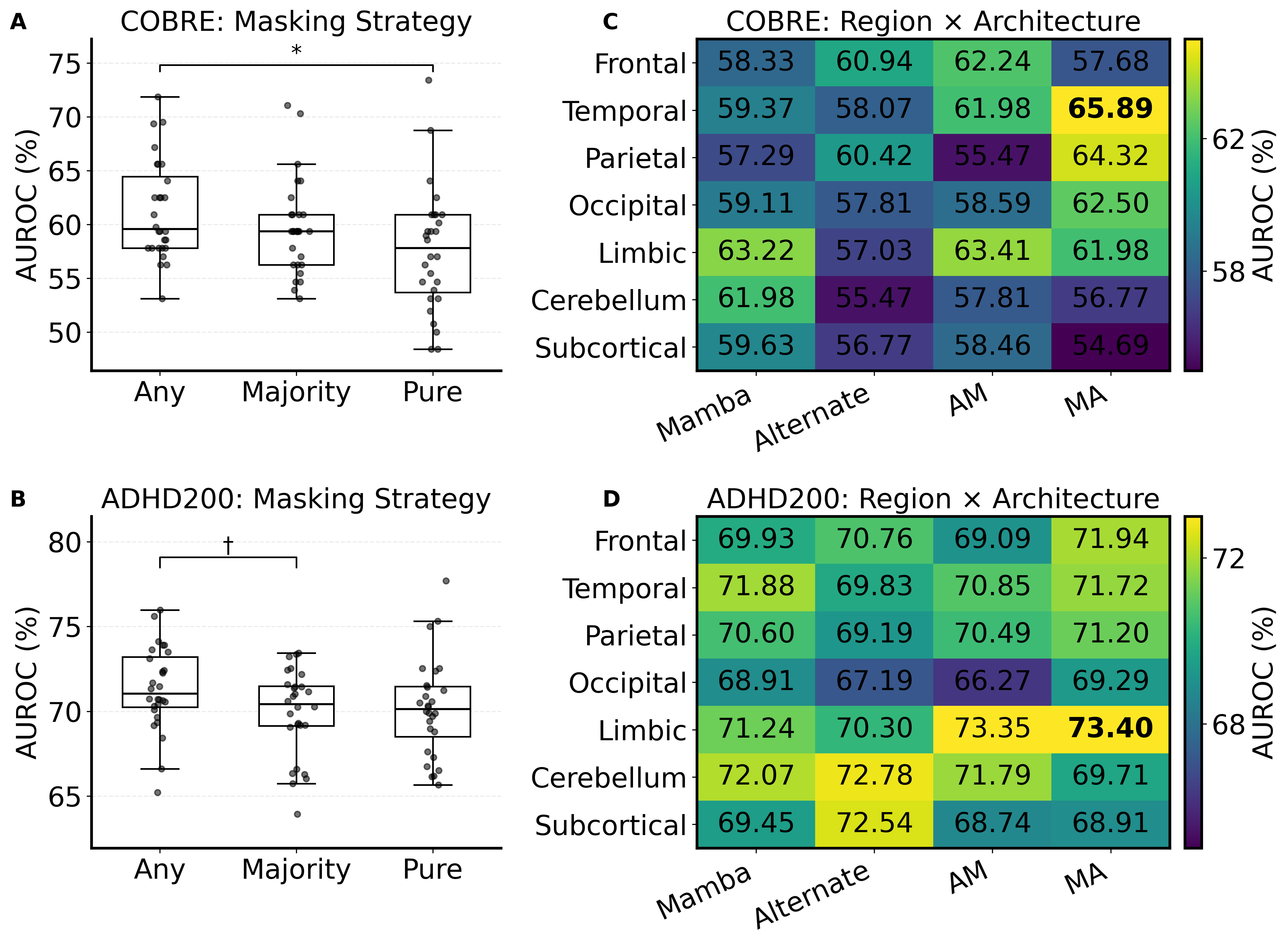}
  \caption{Masking strategy comparison and region-wise architecture performance across datasets.
(A–B) Box plots of ROC–AUC (\%) for three masking strategies (Any, Majority, Pure), aggregated across regions and architectures for (A) COBRE and (B) ADHD-200. Each point represents an individual region–architecture combination. Statistical significance between masking strategies was assessed using the Wilcoxon signed-rank test with Bonferroni correction; significant differences are indicated by asterisk (p < 0.05) and dagger (p < 0.06). (C–D) Heatmaps of mean AUROC (\%) across brain regions (rows) and model architectures (columns) for (C) COBRE and (D) ADHD-200. Each cell reflects performance averaged across masking strategies. Color scales are shown separately for each dataset to enhance within-dataset interpretability.The highest AUROC value across all region–architecture combinations is highlighted.}
  \label{fig:masking_arch_results}
\end{figure}

\section{Results}
\label{headings}
We evaluate \textbf{Rhamba}, a region-aware self-supervised pretraining framework based on hybrid Attention–Mamba architectures, on downstream neuroimaging classification tasks using COBRE (schizophrenia vs.\ control) and ADHD-200 (ADHD vs.\ control). The proposed framework incorporates anatomically guided tokenization and masking strategies, enabling region-aware representation learning from resting-state fMRI. We systematically assess multiple architectural variants, including Mamba-dominant blocks, alternating Attention–Mamba blocks, and asymmetric encoder–decoder combinations, to understand how different modeling strategies impact downstream performance. Evaluation is performed using accuracy (ACC) and area under the receiver operating characteristic curve (AUROC), alongside reconstruction loss to characterize pretraining behavior.

\subsection{Effect of Masking Strategy}
The impact of masking strategy on downstream performance is shown in Figure \ref{fig:masking_arch_results} (A,B). The Friedman test did not reveal a significant overall effect of masking strategy for either dataset (COBRE: $p = 0.111$; ADHD-200: $p = 0.173$), indicating that differences across \textit{Any}, \textit{Majority}, and \textit{Pure} are modest. Pairwise Wilcoxon tests with Bonferroni correction showed selective differences. In COBRE, \textit{Any} outperformed \textit{Pure} ($61.31\%$ vs.\ $57.62\%$, corrected $p = 0.022$), while the difference between \textit{Any} and \textit{Majority} was small and not significant, suggesting comparable performance between these two strategies. In ADHD-200, \textit{Any} achieved higher trend-level significant performance than \textit{Majority} ($71.35\%$ vs.\ $69.92\%$, corrected $p = 0.050$), whereas \textit{Any} and \textit{Pure} showed similar performance. Overall, the influence of masking strategy appears subtle and context-dependent, with no single strategy consistently dominating across datasets.

\subsection{Region-wise Performance Across Architectures}
\label{region-wise performance across architectures}
Region-wise performance across architectures is shown in Figure~\ref{fig:masking_arch_results}(C,D) for COBRE and ADHD-200, respectively. The results reflect how region-specific masking configurations influence optimization on downstream classification tasks. In the COBRE dataset, region-wise averages indicate that Temporal ($61.33 \pm 2.98\%$) and Limbic ($61.41 \pm 2.59\%$) achieve the highest mean performance, while Subcortical ($57.39 \pm 1.86\%$) and Cerebellum ($58.01 \pm 2.44\%$) show comparatively lower values. Across architectures, MA achieves the highest overall performance ($60.55 \pm 3.88\%$), followed by Mamba ($59.85 \pm 1.91\%$) and AM ($59.71 \pm 2.66\%$), with Alternate performing lowest ($58.07 \pm 1.83\%$). The best region–architecture combination is observed for the Temporal region with MA, reaching $65.89\%$ AUROC. 

In the ADHD-200 dataset, the Limbic region yields the highest average performance ($72.08 \pm 1.34\%$), followed by Cerebellum ($71.59 \pm 1.14\%$) and Temporal ($71.07 \pm 0.82\%$), while Occipital shows comparatively lower performance ($67.91 \pm 1.24\%$). Across architectures, MA again achieves the highest mean performance ($70.88 \pm 1.52\%$), outperforming Mamba ($70.58 \pm 1.13\%$), AM ($70.08 \pm 2.13\%$), and Alternate ($70.37 \pm 1.79\%$). The highest AUROC is obtained in the Limbic region with MA ($73.40\%$), closely followed by AM ($73.35\%$). Overall, MA consistently achieves the highest average performance across both datasets, indicating more effective optimization under region-based masking, while region-specific variations highlight task-dependent differences in performance.

\begin{table*}[t]
\centering
\caption{Downstream classification performance using the \textbf{MA} architecture across masking strategies. Values are reported as ACC / AUROC (\%) for each anatomical region in COBRE and ADHD-200. For each dataset, the best Accuracy is highlighted in red and the best AUROC in blue.}
\label{tab:downstream_ma}
\resizebox{\textwidth}{!}{
\begin{tabular}{lccc|ccc}
\toprule
\multirow{2}{*}{Region}
& \multicolumn{3}{c|}{COBRE (ACC / AUROC, \%)}
& \multicolumn{3}{c}{ADHD-200 (ACC / AUROC, \%)} \\
\cmidrule(r){2-4} \cmidrule(l){5-7}
& Any & Majority & Pure & Any & Majority & Pure \\
\midrule
Frontal& 62.50 / 59.77 & 62.50 / 59.38 & 56.25 / 53.91 & 70.21 / 74.11 & 67.02 / 71.45 & 67.20 / 70.26 \\
Temporal & 68.75 / 65.62 & 68.75 / 71.09 & 62.50 / 60.94 & 65.96 / 70.55 & 66.76 / 69.29 & 68.09 / 75.32 \\
Parietal & 62.50 / 65.62 & 53.12 / 53.91 & 68.75 / \textcolor{blue}{73.44} & 67.02 / 70.72 & 69.15 / 71.44 & 69.15 / 71.43 \\
Occipital & 68.75 / 57.81 & \textcolor{red}{75.00} / 70.31 & 56.25 / 59.38 & 63.30 / 70.40 & 68.88 / 69.86 & 65.60 / 67.61 \\
Limbic & 62.50 / 64.06 & 56.25 / 59.38 & 62.50 / 62.50 & 63.30 / 70.08 & 69.41 / 72.44 & 67.02 / \textcolor{blue}{77.69} \\
Cerebellum & 62.50 / 57.03 & 62.50 / 56.25 & 50.00 / 57.03 & 63.83 / 72.33 & \textcolor{red}{71.28} / 71.14 & 64.36 / 65.65 \\
Subcortical & 56.25 / 53.12 & 68.75 / 57.81 & 56.25 / 53.12 & 63.30 / 66.59 & 68.79 / 70.25 & 65.69 / 69.87 \\
\bottomrule
\end{tabular}}
\end{table*}

\subsection{Detailed Analysis of MA Architecture}

Table~\ref{tab:downstream_ma} presents region-wise performance of the MA architecture across masking strategies for both datasets. The results reveal distinct patterns in how masking configurations interact with region-specific optimization. In the COBRE dataset, the highest AUROC is observed in the Parietal under \textit{Pure} masking ($73.44\%$), while the highest accuracy is achieved in the Occipital under \textit{Majority} masking ($75.00\%$). Strong AUROC performance is also observed in the Temporal under \textit{Majority} masking ($71.09\%$), indicating that stricter masking strategies can yield improvements in specific regions. In contrast, \textit{Any} masking provides more consistent but moderate performance across regions.

In the ADHD-200 dataset, the highest AUROC is obtained in the Limbic under \textit{Pure} masking ($77.69\%$), representing the best overall performance across both datasets. The highest accuracy is achieved in the Cerebellum under \textit{Majority} masking ($71.28\%$). High AUROC value is also observed in the Temporal model under \textit{Pure} masking ($75.32\%$), suggesting that stricter, more homogeneous masking can be advantageous in this dataset. Across both datasets, peak performance under the MA architecture varies across masking strategies and regions, with \textit{Majority} and \textit{Pure} frequently achieving the highest values, while \textit{Any} provides more stable but less extreme performance.

For other architectures, including Mamba, Alternate, and AM, performance across masking strategies on the COBRE and ADHD-200 datasets is summarized in Supplementary Tables 2 and 3.

\begin{figure}[!htbp]
  \centering
  \includegraphics[width=1\textwidth]{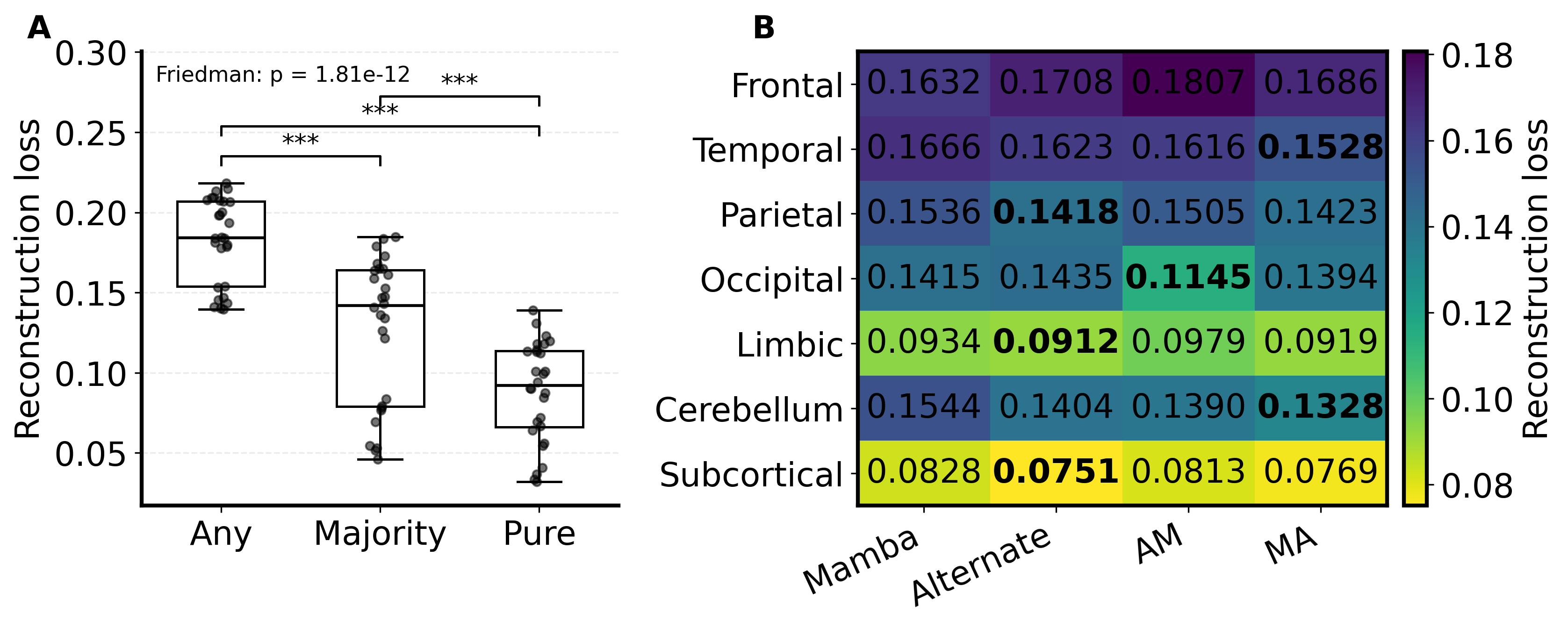}
 \caption{Reconstruction loss across masking strategies, regions, and architectures. 
\textbf{(A)} Box plots showing reconstruction loss for the three masking strategies (\textit{Any}, \textit{Majority}, and \textit{Pure}), aggregated across regions and architectures. Each point represents an individual region--architecture pair. Statistical significance was assessed using the Friedman test, followed by pairwise Wilcoxon signed-rank tests with Bonferroni correction; significant differences are indicated by asterisks (***$p$ < 0.001). 
\textbf{(B)} Heatmap of mean reconstruction loss across brain regions (rows) and model architectures (columns), averaged over masking strategies. Lower values indicate better reconstruction performance. Values corresponding to the lowest reconstruction loss within each region are highlighted, with bold emphasis specifically marking cases where hybrid architectures (\textit{Alternate} or \textit{AM} or \textit{MA}) achieve the minimum, indicating their relative advantage across regions.}
  \label{fig:reconstruction_loss}
\end{figure}

\subsection{Reconstruction Loss Analysis}

Reconstruction loss across masking strategies is shown in Figure~\ref{fig:reconstruction_loss}(A). Masking strategy has a strong effect on reconstruction behavior, as confirmed by the Friedman test ($p = 1.81 \times 10^{-12}$). Pairwise Wilcoxon signed-rank tests with Bonferroni correction indicate significant differences between all masking strategies (\textit{Any} vs.\ \textit{Majority}: $p = 2.23 \times 10^{-8}$; \textit{Any} vs.\ \textit{Pure}: $p = 2.23 \times 10^{-8}$; \textit{Majority} vs.\ \textit{Pure}: $p = 2.69 \times 10^{-5}$). The results show a clear ordering, with \textit{Any} masking producing the highest reconstruction loss ($0.183$), followed by \textit{Majority} ($0.126$), while \textit{Pure} achieves the lowest loss ($0.088$). 

Region-wise reconstruction loss across architectures is shown in Figure~\ref{fig:reconstruction_loss}(B). Reconstruction difficulty varies across regions, with Subcortical and Limbic regions exhibiting lower loss values, while Frontal and Temporal regions show comparatively higher loss. Across architectures, hybrid variants (AM and MA) generally achieve lower reconstruction loss, indicating improved modeling of masked inputs. 

\begin{table}[ht]
\centering
\caption{Performance comparison across COBRE and ADHD datasets (mean $\pm$ std across seeds). Metrics are reported as Accuracy / AUROC (\%). Best Accuracy is highlighted in red and best AUROC in blue.}

\begin{subtable}{\linewidth}
\centering
\caption{COBRE}
\begin{tabular}{l c}
\hline
\textbf{Method} & \textbf{ACC / AUROC} \\
\hline
SwiFT 
& 58.38 $\pm$ 2.99 / 61.30 $\pm$ 4.14 \\

NeuroSTORM (Random + Random) 
& {\color{red}60.66 $\pm$ 2.77} / {\color{blue}62.49 $\pm$ 2.20} \\

NeuroSTORM (Window + Random) 
& 53.80 $\pm$ 7.21 / 59.82 $\pm$ 4.89 \\

NeuroSTORM (Random + Tube) 
& 56.01 $\pm$ 7.17 / 59.88 $\pm$ 7.51 \\

\hline
\end{tabular}
\end{subtable}

\vspace{0.5em}

\begin{subtable}{\textwidth}
\centering
\caption{ADHD}
\begin{tabular}{l c}
\hline
\textbf{Method} & \textbf{ACC / AUROC} \\
\hline
SwiFT 
& \textcolor{red}{66.52 $\pm$ 5.69} / {\color{blue}69.06 $\pm$ 8.32} \\

NeuroSTORM (Random + Random) 
& 60.77 $\pm$ 4.11 / 66.09 $\pm$ 3.11 \\

NeuroSTORM (Window + Random) 
& 64.10 $\pm$ 7.96 / 65.06 $\pm$ 5.94 \\

NeuroSTORM (Random + Tube) 
& 63.13 $\pm$ 3.30 / 67.76 $\pm$ 3.44 \\

\hline
\end{tabular}
\end{subtable}
\label{tab:SOTA}
\end{table}

\subsection{Comparison with Transformer and State-Space Models}
\label{comparison_results}

Table~\ref{tab:SOTA} summarizes the performance of the established volume-based models, SwiFT and NeuroSTORM, under our controlled experimental setting. On COBRE, SwiFT achieved an accuracy of 58.38\% and an AUROC of 61.30\%, whereas the NeuroSTORM variants achieved accuracies ranging from 53.80\% to 60.66\% and AUROCs ranging from 59.82\% to 62.49\%. On ADHD, SwiFT achieved an accuracy of 66.52\% and an AUROC of 69.06\%, while the NeuroSTORM variants achieved accuracies ranging from 60.77\% to 64.10\% and AUROCs ranging from 65.06\% to 67.76\%.

For reference, the selected region-aware Rhamba-MA configurations also showed competitive performance across the same five independent repeats. Under Majority masking on COBRE, the Temporal configuration achieved 68.13 $\pm$ 1.40\% accuracy and 67.81 $\pm$ 4.76\% AUROC, whereas the Occipital configuration achieved 71.32 $\pm$ 3.42\% accuracy and 68.44 $\pm$ 1.71\% AUROC. On ADHD-200, the Limbic configuration achieved 65.99 $\pm$ 2.13\% accuracy and 71.44 $\pm$ 1.00\% AUROC, while the Cerebellum configuration achieved 66.01 $\pm$ 3.19\% accuracy and 70.38 $\pm$ 1.67\% AUROC. These values are reported to characterize the behavior of representative region-aware configurations rather than to establish a definitive ranking relative to SwiFT or NeuroSTORM. Given the large number of Rhamba variants evaluated, we avoid emphasizing a single best-performing configuration to reduce the risk of a multiple-comparison winner's curse.

\subsection{Interpretability Analysis}
\begin{figure}[!htbp]
  \centering
  \includegraphics[width=1\linewidth]{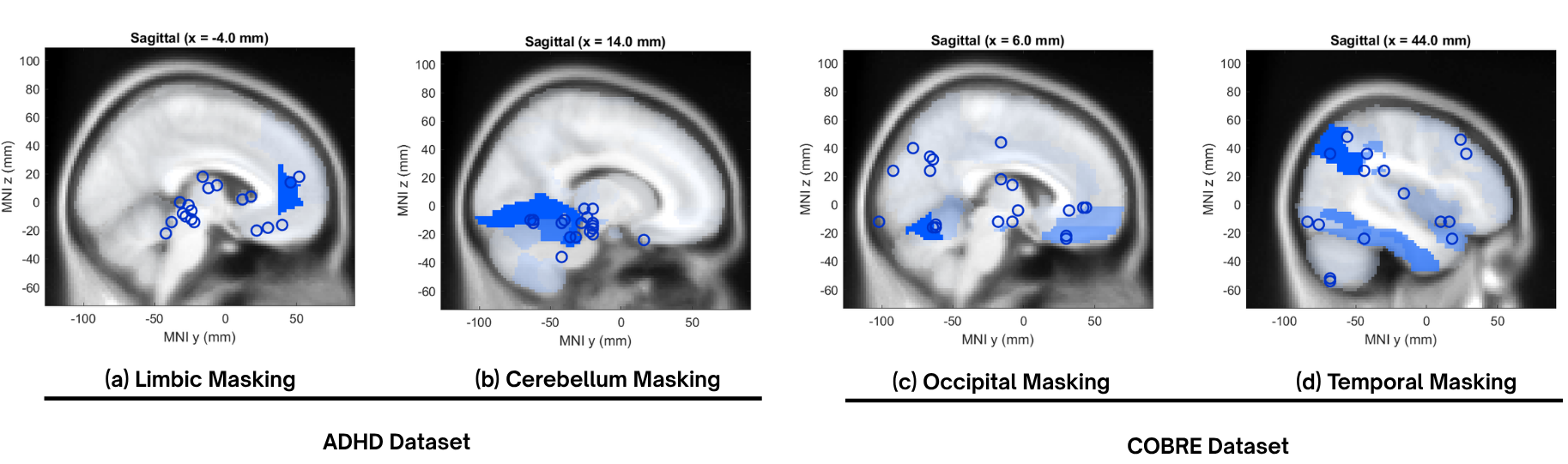}
  \caption{Interpretation maps generated using the Integrated Gradients (IG) method shown in sagittal view: (a) limbic region masking and (b) cerebellum masking for the ADHD dataset; (c) occipital masking and (d) temporal region masking for the COBRE dataset. Circles highlight the top 20 contributing regions of interest (ROIs), while the colored maps represent attribution values, with light-to-dark intensity indicating increasing contribution.}
  \label{fig:interpretation}
\end{figure}

Regional mean attribution values from the region-aware self-supervised resting-state fMRI model were projected onto the AAL3 atlas in MNI152 2 mm space. Each atlas-defined ROI was assigned its corresponding mean attribution value, producing a parcelwise attribution map for visualization. This map reflects model-derived regional importance rather than voxelwise statistical activation. 

Figure ~\ref{fig:interpretation} show attribution maps generated to improve anatomical interpretation. These maps were generated based on the ROI’s AAL3 atlas labels and their corresponding mean attribution value; ROIs with less than 10 voxels were excluded from the map. The top 20 ROIs are circled in blue, representing the higher contributions to classify patients versus healthy controls. Supplementary Figures S1–S4 present sagittal, coronal, and axial views.

In the COBRE dataset, temporal and occipital masking strategies reveal regions that are most consistent with a network profile spanning internal mentation and memory (DMN-like regions), value and affective evaluation (vmPFC/OFC/insula/pallidum), executive-attentional control (middle frontal, cingulate, parietal, Crus II), and high-level visual association (occipital-temporal-fusiform regions). Neuroimaging characterization of schizophrenia highlights altered or mostly hyperconnectivity in the default mode network (DMN), reduced prefrontal connectivity, and altered thalamocortical/frontolimbic/cortico-cerebellar networks \cite{Sasabayashi2023,Karbasforoushan2012}. A previous study investigated the voxel-wise resting-state functional connectivity of the thalamic nucleus in drug-naïve first-episode schizophrenia with auditory verbal hallucinations (AVH), linking this positive symptom with the PuM nucleus in the thalamus \cite{Wei2022}. 

Model’s behavior with the ADHD-200 dataset implicates regions in the default mode network (DMN), and regions with broad and overlapping functions spanning visual association, memory, affective, interoceptive, and higher-order cognitive systems. A meta-analysis of 20 seed-based resting-state studies (944 ADHD patients vs. 1,121 controls) found reduced connectivity within the core DMN (centered on the posterior cingulate cortex) but elevated connectivity in the dorsal medial prefrontal cortex subsystem, and further found reduced connectivity between the DMN and cognitive control, salience, and affective/motivational networks \cite{Sutcubasi2020}. Atypical DMN connectivity in children may also reflect a delayed or disrupted maturational process. Fair et al. (Biological Psychiatry, 2010) found that the developmental consolidation of the default network seen in typically developing children does not proceed normally in those diagnosed with ADHD \cite{Fair2010}.Overall, ADHD involves disrupted coordination among at least three interacting networks; the frontoparietal executive control network, the subcortical (striato-thalamic) circuit, and the DMN; with the cerebellum increasingly implicated as a fourth node.
\section{Ablation Analysis}

\begin{figure}[t]
  \centering
  \includegraphics[width=1.\linewidth]{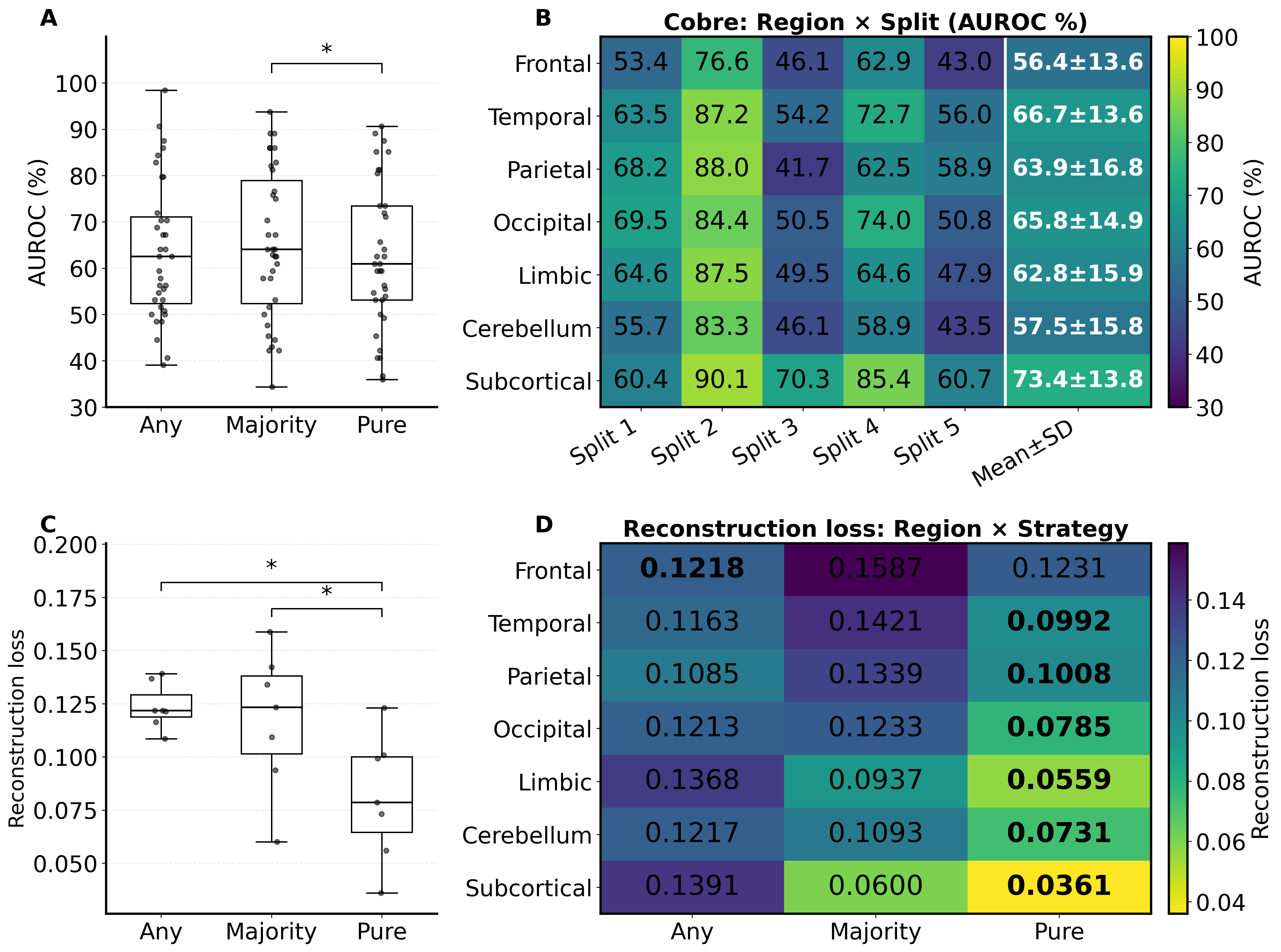}
  \caption{Masking strategy comparison in COBRE and reconstruction loss comparison under a fixed mask budget. (A) Box plots of ROC–AUC (\%) for three masking strategies (\textit{Any}, \textit{Majority}, \textit{Pure}) with a fixed number of masked patches, using the minimum Pure mask budget, aggregated across regions and five different splits for COBRE. Each point represents an individual model trained on the MA architecture. The averaged AUROC value across all regions is highlighted. (B) Heatmap of mean AUROC (\%) across brain regions (rows) and split (columns) for COBRE. (C) Box plots showing reconstruction loss for the three masking strategies (\textit{Any}, \textit{Majority}, and \textit{Pure}) under a fixed masking budget for each region. (D) Heatmap of mean reconstruction loss across brain regions (rows) and masking strategy (columns). Lower values indicate better reconstruction performance. Values corresponding to the lowest reconstruction loss within each region are highlighted in bold. Statistical significance between masking strategies was assessed using the Wilcoxon signed-rank test. Significant differences are indicated by an asterisk (p<0.05).}
  \label{fig:masking_arch_results_ablation}
\end{figure}

\subsection{Matched-Budget Masking Effects on Downstream Performance and Reconstruction}
To control for the confounding effect of the number of masked patches, we repeated the masking-strategy analysis under a fixed mask budget, where the number of masked patches in each region was set to the minimum Pure mask count, as shown in Fig.~\ref{fig:masking_arch_results_ablation}(A--B). Under this matched setting, the overall differences among \textit{Any}, \textit{Majority}, and \textit{Pure} were modest (COBRE: Friedman $p = 0.130$), with only the \textit{Majority} vs.\ \textit{Pure} comparison showing a trend-level difference before correction (uncorrected $p = 0.0267$, Bonferroni-corrected $p = 0.0801$). The corresponding mean AUROC values were 63.64\% for \textit{Any}, 65.08\% for \textit{Majority}, and 62.63\% for \textit{Pure}, indicating that mask budget explains much of the variation observed in the original analysis. Region-wise results under the same fixed-budget setting showed a similar pattern: in COBRE, \textit{Temporal} remained among the strongest regions after \textit{Subcortical}, consistent with the earlier results.

Reconstruction loss across masking strategies is shown in Fig.~\ref{fig:masking_arch_results_ablation}(C--D). Masking strategy had a significant effect on reconstruction behavior, as confirmed by the Friedman test ( $p = 0.0183$). The mean reconstruction loss was highest for \textit{Any} (0.1236 $\pm$ 0.0109), followed by \textit{Majority} (0.1173 $\pm$ 0.0330), and lowest for \textit{Pure} (0.0810 $\pm$ 0.0295). Pairwise Wilcoxon signed-rank tests showed no significant difference between \textit{Any} and \textit{Majority} (uncorrected $p = 0.9375$, Bonferroni-corrected $p = 1$), whereas \textit{Majority} vs.\ \textit{Pure} remained significant after correction (uncorrected $p = 0.0156$, Bonferroni-corrected $p = 0.0469$). The \textit{Any} vs.\ \textit{Pure} comparison showed the same direction but did not survive Bonferroni correction (uncorrected $p = 0.0313$, Bonferroni-corrected $p = 0.0938$). Region-wise reconstruction loss across masking strategies is shown in Figure~\ref{fig:masking_arch_results_ablation}(D). Reconstruction difficulty varies across regions, with Subcortical and Limbic regions exhibiting lower loss values, while Frontal and Temporal regions show comparatively higher loss. A possible rationale is that these regions differ in their masked budgets, as also observed in the previous experiment in Fig.~\ref{fig:reconstruction_loss}.

\begin{figure}[t]
  \centering
  \includegraphics[width=1.\linewidth]{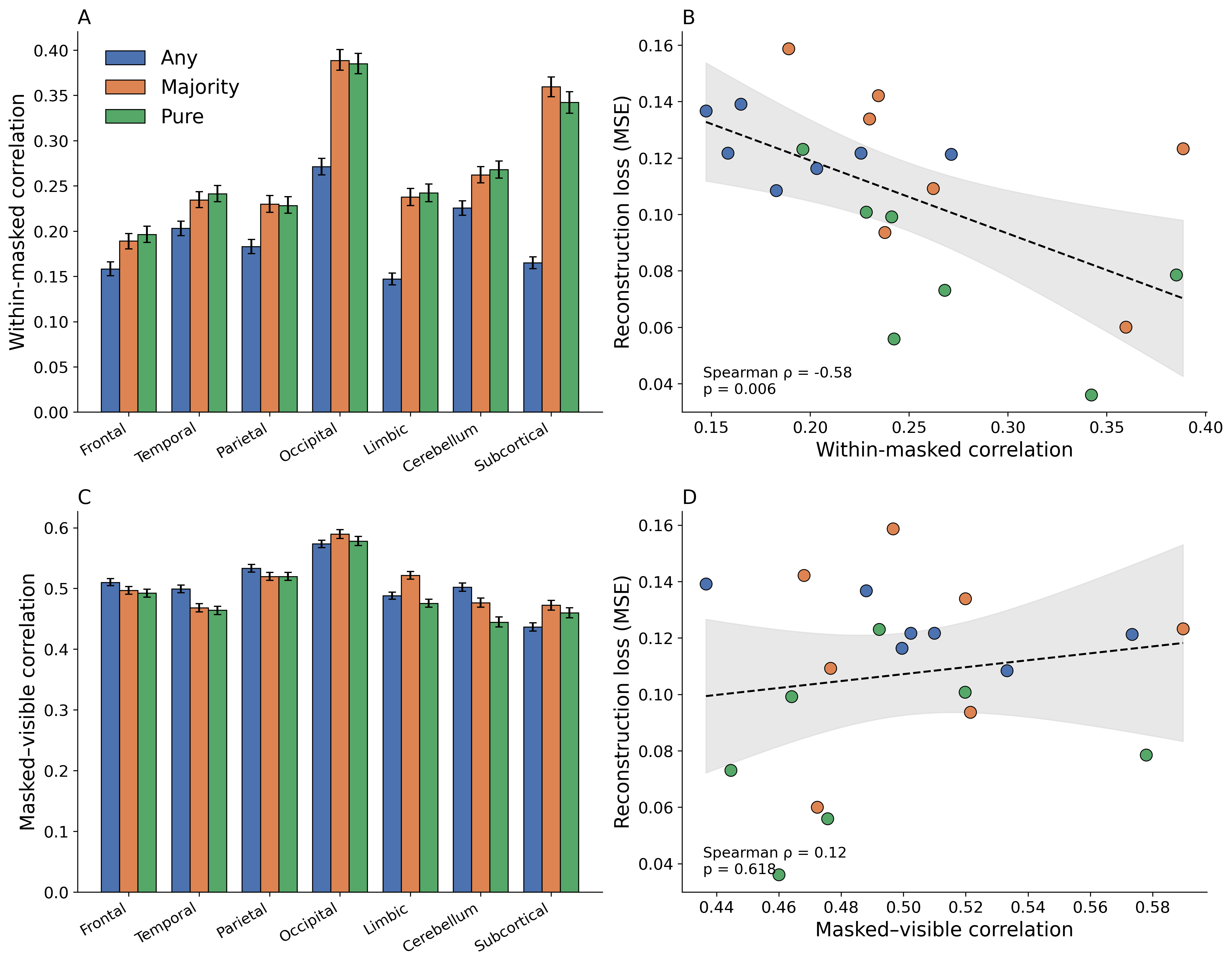}
  \caption{Within-mask and masked–visible correlations of masked patch sets in relation to reconstruction loss across masking configurations. Model-free BOLD statistics computed from 500 ABIDE subjects, with masked-set size matched across configurations within each region. (A) Within-mask correlation (mean pairwise correlation among masked patches) per region and configuration; bars show the mean and error bars the bootstrap 95\% CI (colors: Any, Majority, Pure). (B) Reconstruction loss versus within-mask correlation across the 21 region × configuration points; each point is one region × configuration, colored by configuration. Configurations that mask internally redundant patch sets are reconstructed with lower loss (Spearman $\rho = -0.58,\; p = 0.006$). (C) Masked–visible correlation (mean correlation of masked patches with their six nearest visible neighbors) per region and configuration. (D) Reconstruction loss versus masked–visible correlation; this proximity metric is unrelated to reconstruction loss (Spearman $\rho = +0.12,\; p = 0.618$).}
  \label{fig:confi_correlation}
\end{figure}

Reconstruction loss differed systematically across masking configurations: averaged over regions, Pure masking was easiest (MSE =0.081), followed by Majority (0.117) and Any (0.124), so Pure masking reduced reconstruction error by roughly 35\% relative to Any. Across these controlled experiments, we observed a pattern consistent with the main analysis: the masking strategy strongly influenced reconstruction behavior. Reconstruction loss followed the general ordering (\textit{Any} > \textit{Majority} > \textit{Pure}). The \textit{Majority} strategy exhibited the widest variability, spanning the lowest to the highest reconstruction losses, whereas \textit{Any} remained consistently high with a smaller standard deviation. However, these reconstruction-loss patterns did not directly translate into downstream performance, as also observed in our main analysis.

To identify the source of this difference, we asked which model-free property of the masked patch set predicts reconstruction loss across the 21 region × configuration conditions. Reconstruction difficulty tracked the intrinsic redundancy of the masked set. Configurations whose masked patches were more mutually correlated were reconstructed with lower loss (within-mask correlation vs. loss: Spearman $\rho = -0.58,\; p = 0.006$; Fig. \ref{fig:confi_correlation}A–B). Consistent with this, the more restrictive configurations masked more internally coherent tissue: mean within-mask correlation rose from 0.19 (Any) to 0.27 (Majority and Pure).
Masked–visible correlation was essentially constant across configurations (mean 0.51, 0.51, and 0.49 for Any, Majority, Pure) and was uncorrelated with reconstruction loss (Spearman $\rho = +.12, p = .618$; Fig. \ref{fig:confi_correlation}C–D). Neither masked-patch signal variance ($\rho=-0.15, p=0.51$) explained the effect, while the spatial compactness of the masked set showed only a weaker, non-significant trend by rank correlation (mean pairwise distance: $\rho=+0.34, p=0.11$).
Together, these results suggest that reconstruction loss depends on how similar the masked patches are to one another: groups of patches that are more similar and compact are easier to reconstruct than more mixed groups.

Table~\ref{tab:split_comparison} summarizes the held-out-split AUROC performance of SwiFT and NeuroSTORM on COBRE. Across the five splits, SwiFT achieved a mean AUROC of 64.52\%, while the NeuroSTORM variants ranged from 61.56\% to 63.91\%. Under the same evaluation setting, the Rhamba-MA configurations showed competitive performance, with the Temporal configuration achieving an AUROC of 66.7\% and the Subcortical configuration achieving 73.4\%. These results indicate that region-aware masking can yield competitive performance relative to established volume-based fMRI models, while the substantial variation across held-out splits further supports a cautious interpretation of absolute model rankings.

\begin{table}[t]
\centering
\caption{Cobre comparison with baseline models under the held-out-split evaluation setting. Reported values are AUROC only (\%). The final column shows mean $\pm$ standard deviation across the five held-out splits.}
\label{tab:split_comparison}
\resizebox{\textwidth}{!}{
\begin{tabular}{lcccccc}
\toprule
\textbf{Method} & \textbf{Split 1} & \textbf{Split 2} & \textbf{Split 3} & \textbf{Split 4} & \textbf{Split 5} & \textbf{Mean $\pm$ SD} \\
\midrule
SwiFT
& 54.70 & 83.60 & 53.50 & 77.30 & 53.50
& 64.52 $\pm$ 14.72 \\

NeuroSTORM (Random + Random)
& 64.06 & 76.56 & 56.25 & 59.38 & 56.25
& 62.50 $\pm$ 8.48 \\

NeuroSTORM (Window + Random)
& 54.69 & 82.03 & 56.25 & 63.28 & 51.56
& 61.56 $\pm$ 12.22 \\

NeuroSTORM (Random + Tube)
& 62.50 & 74.22 & 65.62 & 62.50 & 54.69
& 63.91 $\pm$ 7.04 \\
\bottomrule
\end{tabular}}
\end{table}

\begin{figure}[t]
  \centering
  \includegraphics[width=1.\linewidth]{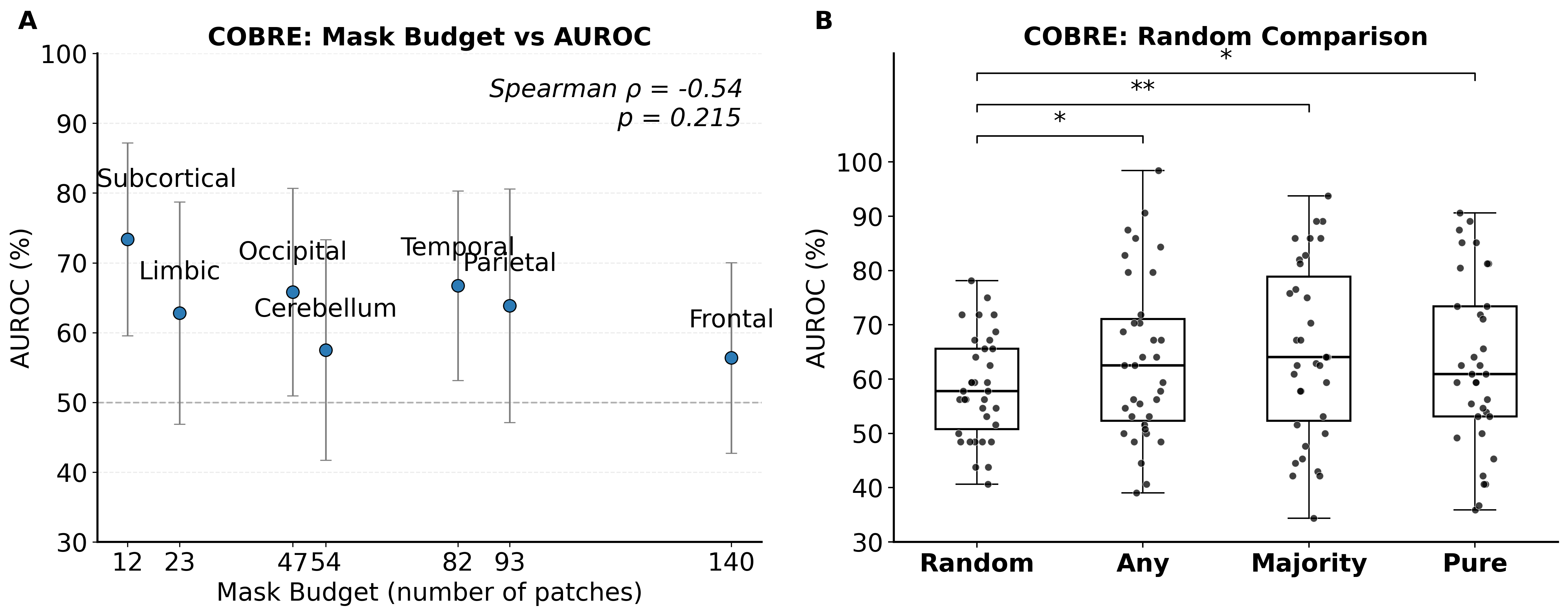}
  \caption{Mask budget dependence and comparison with a random baseline in COBRE. (A) Scatter plot showing the relationship between the number of masked patches and downstream AUROC (\%) across brain regions under a fixed-budget setting. Each point represents one region-specific masking configuration, and the Spearman correlation is reported in the panel. (B) Box plots of AUROC (\%) comparing region-aware masking strategies (\textit{Any}, \textit{Majority}, \textit{Pure}) against a random baseline with the same mask budget. Each point represents an individual model trained on the MA architecture. Statistical significance between groups was assessed using the Wilcoxon signed-rank test, with significant differences indicated by asterisks.}
  \label{fig:mask_budget_random}
\end{figure}

\subsection{Mask Budget and Random Baseline Comparison}
To examine whether downstream performance is driven primarily by the number of masked patches, we compared the regional mask budget, that is, the total number of patches masked within each region, against AUROC in COBRE (Fig.~\ref{fig:mask_budget_random}(A)). Although regions with smaller budgets such as \textit{Subcortical} and \textit{Limbic} achieved higher AUROC than some higher-budget regions, the overall association was not significant (Spearman $\rho = -0.54$, $p = 0.215$), indicating that mask budget alone does not explain the observed performance pattern. 

We next compared region-aware masking against a random baseline in which the same number of patches was masked, but without anatomical guidance (Fig.~\ref{fig:mask_budget_random}(B)). \textit{Any} and \textit{Majority} significantly outperformed \textit{Random} after Bonferroni correction (Random vs.\ \textit{Any}: corrected $p = 0.030$; Random vs.\ \textit{Majority}: corrected $p = 0.014$), whereas \textit{Pure} showed the same favorable direction but did not survive correction (corrected $p = 0.167$). Overall, these results suggest that region-aware masking provides benefits beyond those of random selection within the same mask budget. We also included Motor Neuron Disease (MND) as a third downstream disease dataset, with results reported in Supplementary Figs. 5 and 6. Consistent with the findings on the COBRE dataset, the results on the MND dataset show the same trend: random initialization consistently yields lower performance than the evaluated strategies.

\begin{figure}[t]
  \centering
  \includegraphics[width=1.\linewidth]{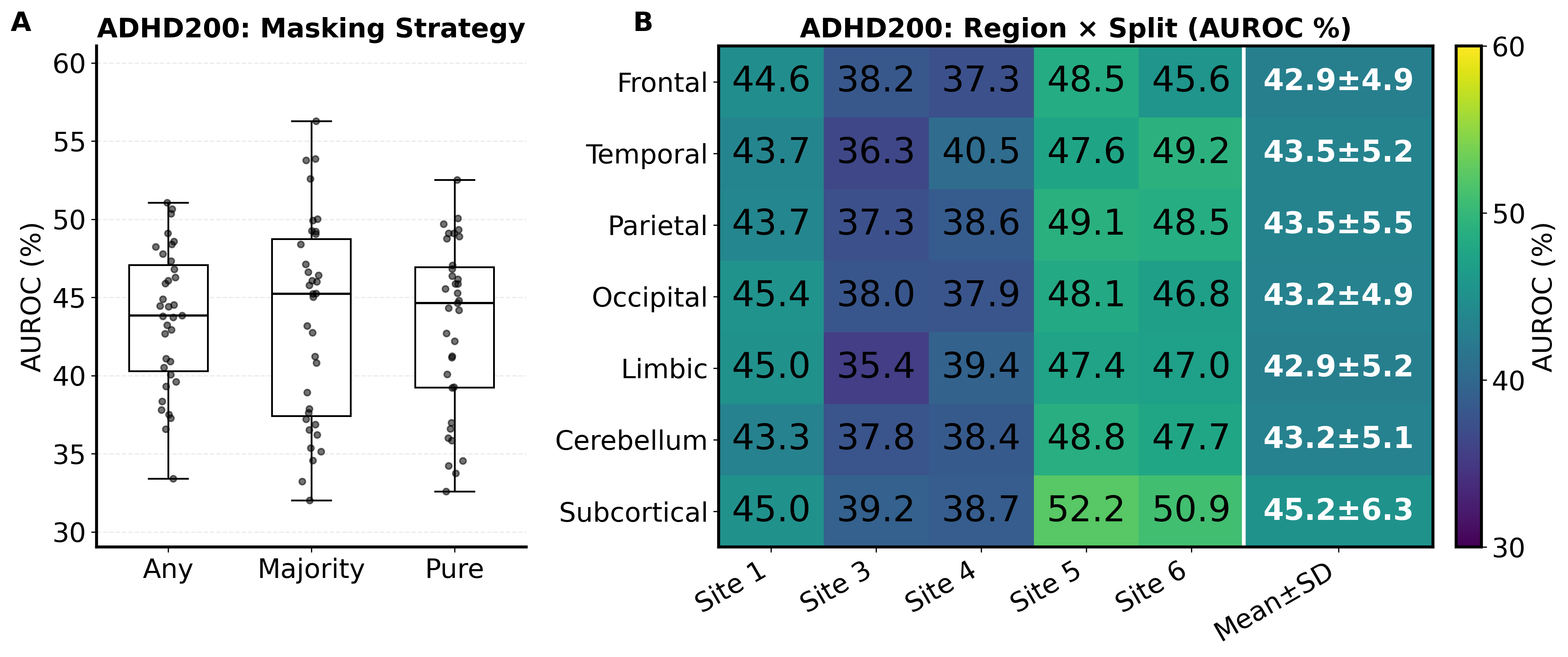}
  \caption{ADHD-200 site-generalization results under a held-out-site setting. (A) Box plots of ROC–AUC (\%) for three masking strategies (\textit{Any}, \textit{Majority}, \textit{Pure}) across the five held-out sites, using the MA architecture. Each point represents an individual model trained on one site split. (B) Heatmap of mean AUROC (\%) across brain regions (rows) and held-out sites (columns), with the corresponding mean $\pm$ SD shown in the final column. The results indicate limited cross-site generalization in ADHD-200, with all strategies remaining near or below chance-level performance.}
  \label{fig:adhd_site_results}
\end{figure}

\begin{table}[t]
\centering
\caption{ADHD-200 comparison with baseline models under the held-out-site evaluation setting. Reported values are AUROC (\%). The final column shows mean $\pm$ standard deviation across the five held-out sites.}
\label{tab:adhd200_baselines_auroc}
\resizebox{\textwidth}{!}{
\begin{tabular}{lccccc c}
\hline
\textbf{Method} & \textbf{Site 1} & \textbf{Site 3} & \textbf{Site 4} & \textbf{Site 5} & \textbf{Site 6} & \textbf{Mean $\pm$ SD} \\
\hline
SwiFT
& 43.48 & 45.43 & 43.88 & 48.50 & 42.46
& 44.75 $\pm$ 2.35 \\

NeuroSTORM (Random + Random) & 49.13 & 50.33 & 52.14 & 54.63 & 47.87 & 50.82 $\pm$ 2.65 \\ 
NeuroSTORM (Random + Tube) & 47.63 & 43.88 & 50.92 & 53.30 & 46.93 & 48.53 $\pm$ 3.66 \\ 
NeuroSTORM (Window + Random) & 46.13 & 41.57 & 48.69 & 50.21 & 51.77 & 47.67 $\pm$ 4.00 \\
\hline
\end{tabular}}
\end{table}

\subsection{ADHD-200 Site-Generalization Bottleneck}
The ADHD-200 results highlight a clear site-generalization bottleneck. Under the held-out-site evaluation in Fig.~\ref{fig:adhd_site_results}(A--B), all masking strategies remained below chance-level performance, with only modest variation across \textit{Any}, \textit{Majority}, and \textit{Pure}. This indicates that, unlike COBRE, the dominant challenge in ADHD-200 is not masking strategy but the strong heterogeneity across acquisition sites. In contrast, the earlier mixed-split results are shown in Fig.~\ref{fig:masking_arch_results} for a setting in which training and test subjects share the same overall site distribution, and therefore should be interpreted as a less stringent estimate of generalization. The baseline methods reported in Table~\ref{tab:adhd200_baselines_auroc} also demonstrate the difficulty of the held-out-site setting. SwiFT remained below chance-level AUROC across sites, whereas NeuroSTORM (Random + Random) reached approximately chance-level performance on average (50.82 $\pm$ 2.65\%) but did not show robust above-chance generalization. The other NeuroSTORM variants also remained below chance on average. Overall, these findings suggest that site shift remains a major bottleneck for ADHD-200, limiting the extent to which masking design or backbone architecture alone can improve downstream classification under strict site-generalization settings.

\begin{figure}[t]
  \centering
  \includegraphics[width=1.\linewidth]{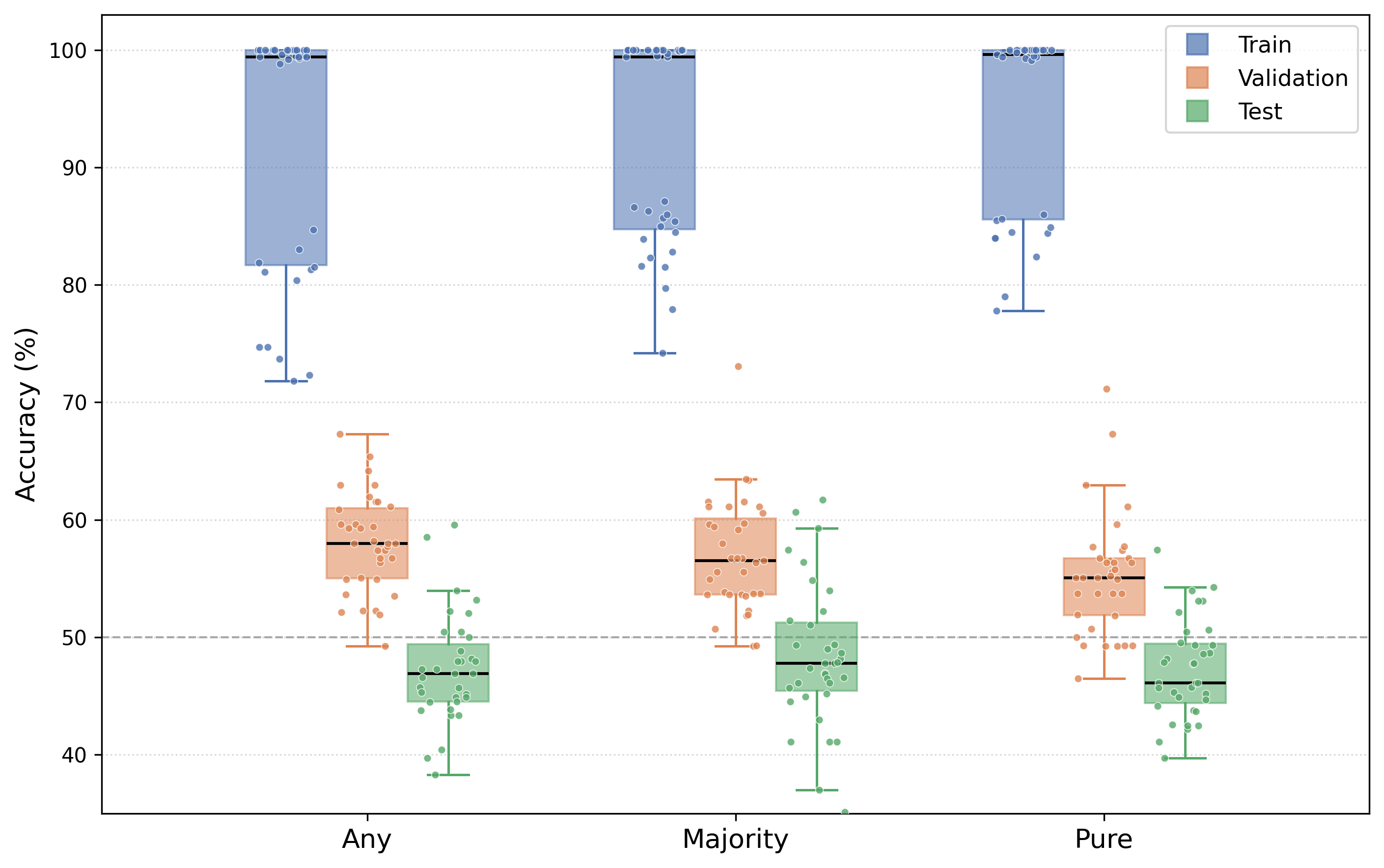}
  \caption{Generalization gap between training, validation, and cross-site test accuracy in ADHD200. For each label configuration (Any, Majority, Pure), box plots show the distribution of best-checkpoint accuracy on the training, validation, and held-out test sets, aggregated across all seven anatomical regions and five cross-site splits (n = 105 runs per split type). Boxes span the interquartile range with the median line; whiskers extend to 1.5× IQR and individual runs are overlaid as points. The dashed line marks chance level (50\%).}
  \label{fig:adhd_generalization_gap}
\end{figure}

We further examined per-split accuracy on the training, validation, and test partitions to characterize the model's generalization behavior in Fig. \ref{fig:adhd_generalization_gap}. Training accuracy was near ceiling (mean 93.1\%, median 99.5\%, SD 9.1\%), indicating that the model fit the training data. Accuracy dropped markedly on the validation set used for model selection (mean 56.7\%, median 56.5\%, SD 4.8\%) and declined further on the held-out cross-site test set (mean 47.5\%, median 46.9\%, SD 5.3\%), falling to around chance level. This pattern was consistent across all three label configurations (test accuracy: Any 46.9 ± 5.4\%, Majority 48.3 ± 6.1\%, Pure 47.3 ± 4.1\%; mean ± SD). Together, these results reveal a substantial generalization gap: the model's strong in-distribution fit does not transfer to unseen sites, with test performance dropping roughly 9 percentage points below validation and approaching chance under cross-site evaluation.

\section{Discussion}

In this work, we introduce \textbf{Rhamba}, a region-aware hybrid Attention–Mamba architecture, and examine the role of anatomically guided masking strategies during self-supervised pretraining in shaping the quality of representations learned from large-scale resting state fMRI data. Models are pretrained on the ABIDE dataset and subsequently fine-tuned on COBRE and ADHD-200 for schizophrenia and attention-deficit/hyperactivity disorder classification, using multiple architectural variants, including Mamba, Alternate, Attention–Mamba (AM), and Mamba–Attention (MA). By integrating architectural design with anatomically informed masking strategies, this study provides a systematic analysis of the impact of pretraining constraints on downstream optimization and generalization in neuroimaging models.

\subsection{Computational Perspective}
A central component of this framework is the use of patch-based embeddings that are aligned with anatomically defined regions of interest, enabling region-aware masking during self-supervised pretraining. Building on this representation, we employ three masking strategies, \emph{Any}, \emph{Majority}, and \emph{Pure}, which impose progressively stricter constraints on patch selection during reconstruction. These strategies directly regulate the balance between spatial heterogeneity and homogeneity in the pretraining objective. The \emph{Any} strategy introduces substantial boundary mixing and increases the number of masked patches, exposing the model to diverse and distributed spatial signals. In contrast, \emph{Pure} masking restricts reconstruction to spatially homogeneous patches, simplifying the learning objective, while \emph{Majority} provides an intermediate regime that balances dominant regional structure with residual contextual information.

The region-aware masking design in Rhamba leads to a clear separation between reconstruction behavior and downstream performance. Reconstruction loss follows a consistent ordering (\emph{Any} $>$ \emph{Majority} $>$ \emph{Pure}), indicating that heterogeneous masking, together with an increased number of masked patches, substantially increases reconstruction difficulty. However, this ordering does not translate directly to downstream performance. In ADHD-200, \emph{Any} and \emph{Pure} achieve comparable results, while in COBRE, \emph{Any} and \emph{Majority} show similar performance. These findings indicate that reconstruction difficulty is not a reliable proxy for representation quality, consistent with prior observations in masked autoencoding and contrastive learning that representation quality is not directly determined by reconstruction fidelity or task difficulty \cite{chen2020simple,he2022masked}. Moreover, reconstruction-based objectives tend to prioritize dominant signal components that are not necessarily informative for downstream tasks \cite{balestriero2024learning}. Instead, the structure of the masking strategy plays a more critical role in shaping learned representations. The \emph{Pure} strategy, which restricts masking to spatially homogeneous regions, emphasizes local feature reconstruction and reduces the complexity of the learning objective. This behavior aligns with findings that local context is often sufficient for reconstruction tasks, leading models to rely on short range dependencies rather than learning richer global representations \cite{chen2025local}. In contrast, the \emph{Any} strategy introduces heterogeneous masking patterns and increases the number of masked patches, effectively acting as a structured perturbation that encourages the model to integrate information across region boundaries. Such masking induced perturbations can be interpreted as implicit augmentation, promoting alignment of semantically related features and improving representation transferability, consistent with principles established in contrastive learning and masked modeling \cite{chen2020simple,zhang2022mask}. Together, these results suggest that the effectiveness of region-aware masking arises not from reconstruction difficulty alone, but from its ability to control the balance between local consistency and cross-regional interaction, thereby guiding the model toward more informative and transferable representations. Importantly, the relative benefit of these masking strategies appears to depend on the nature of the downstream task. Brain disorders can manifest through varying patterns of functional organization, ranging from distributed network-level alterations to more localized disruptions in brain activity \cite{menon2011large, van2014brain}. In settings where dysfunction is distributed across large-scale functional systems, masking strategies that encourage integration of information across regions may offer advantages. In contrast, when alterations are more spatially constrained, representations that preserve intra-regional consistency may be more effective. This perspective underscores the importance of aligning self-supervised pretraining strategies with the underlying characteristics of brain organization relevant to the downstream task.

The matched-budget ablation further clarifies the role of masking strategy in Rhamba. In the original analysis, Any, Majority, and Pure differed not only in anatomical specificity but also in the number of masked patches. After controlling this factor by matching the mask budget within each region, downstream differences among masking strategies became more modest, suggesting that part of the apparent performance variation was attributable to differences in mask size rather than anatomical specificity alone. However, reconstruction loss continued to show a structured ordering, with Pure remaining easier to reconstruct than more heterogeneous configurations. This indicates that reconstruction difficulty is not governed solely by the amount of missing information, but also by the intrinsic organization of the masked target set. The model-free analysis supports this interpretation: masked configurations with higher within-set BOLD correlation were reconstructed with lower loss, whereas correlation between masked patches and nearby visible patches was not predictive. Thus, the ease of reconstruction appears to depend more on the internal redundancy and coherence of the masked patches than on their proximity to visible context. These findings refine our interpretation of region-aware masking: anatomical constraints influence the statistical structure of the reconstruction target, but lower reconstruction loss does not necessarily imply more transferable downstream representations.

Architectural design plays a critical role in determining representation quality. Across both datasets, the Mamba–Attention (MA) configuration consistently achieves the strongest average performance, outperforming the Mamba-dominant, Attention–Mamba (AM), and Alternate variants. This indicates that the relative placement and interaction between attention and state space components are central to effective representation learning. Mamba layers efficiently model long-range temporal dependencies with linear complexity \cite{gu2022efficient, gu2023mamba}, but they do not provide the same explicit pairwise global interaction mechanism as self-attention. In contrast, attention mechanisms explicitly model global token interactions \cite{vaswani2017attention}, but their computational cost increases substantially with sequence length. In our implementation, the MA configuration should be interpreted as Mamba-dominant: Mamba blocks form the primary encoding pathway, while the final encoder stage includes full self-attention before the latent representation, and the decoder is attention-based. This design preserves the efficiency of state-space modeling in earlier hierarchical stages while enabling global contextual integration at the deepest encoder stage, where the reduced token sequence makes attention more computationally tractable. Because the decoder is discarded during downstream fine-tuning, the downstream benefit of MA should be understood primarily as arising from the Mamba-dominant encoder shaped by final-stage attention and reconstruction supervision, rather than from direct use of the decoder at inference. This design is consistent with recent advances in hybrid sequence modeling, where combining attention and state space mechanisms can improve performance and scalability compared with relying on either mechanism alone \cite{lieber2024jambahybridtransformermambalanguage, kannan2025brainmt}. In resting state fMRI, where brain activity reflects both distributed interactions and extended temporal dynamics, such hybrid architectures provide a more suitable inductive bias for balancing global contextual integration with efficient sequence modeling. The consistent performance of MA across datasets highlights the importance of jointly modeling global context and long-range temporal structure, while also suggesting that the placement of attention within a predominantly state-space backbone can substantially influence representation quality. 

Region-wise analysis indicates that performance is governed by the interaction between masking strategy and architectural design rather than by fixed regional dominance. Different regions achieve peak performance under different masking configurations, and these patterns vary across datasets. \emph{Any} and \emph{Majority} provide more stable performance across regions, whereas \emph{Pure} yields peak performance in specific cases but lacks consistency. This behavior suggests that region aware masking acts as an optimization constraint that shapes the diversity and structure of learned representations, rather than enforcing strictly localized learning. From a modeling perspective, these findings emphasize the importance of balancing structural priors with signal diversity. While stricter masking simplifies reconstruction, incorporating heterogeneous spatial context through partial overlap (\emph{Any}) or dominant occupancy (\emph{Majority}) leads to more robust and generalizable representations.

The variability observed across datasets and region-aware configurations suggests that performance is shaped by dataset-specific factors, including sample size, acquisition heterogeneity, site distribution, class balance, and anatomical alignment quality. Accordingly, the comparison with SwiFT and NeuroSTORM should be interpreted as evidence that Rhamba achieves competitive performance relative to established volume-based fMRI models under the same experimental setting, rather than as a definitive claim of universal superiority. Because Rhamba includes multiple variants across anatomical regions, masking strategies, and architectural configurations, we avoid emphasizing a single best-performing configuration as universally optimal. Instead, our results suggest that region-aware masking provides a flexible and transferable design principle for introducing structured spatial perturbations, whose effectiveness depends on the interaction between anatomical priors, model architecture, and dataset characteristics. This perspective further suggests that anatomically structured masking could be integrated with other transformer-based or state-space backbones to test whether region-aware pretraining improves representation learning independently of the specific sequence model.

\subsection{Neuroscientific Interpretation}
Region-aware masking provides a structured framework for examining how spatial constraints influence representation learning across different data conditions. In this setting, input volumes are first transformed into patch-level embeddings and aligned with anatomically defined regions of interest, allowing region-specific masking to act as a controlled perturbation of spatial information. Consequently, observed performance patterns reflect how region-constrained representations affect optimization, rather than direct evidence of localized anatomical importance. In the COBRE dataset, which exhibits relatively consistent anatomical alignment, the Temporal and Occipital regions show stable and competitive performance across masking strategies (Section~\ref{region-wise performance across architectures}). This consistency suggests that representations associated with these regions capture reproducible patterns relevant for downstream classification. Prior neuroimaging studies have reported altered functional activity and connectivity in both temporal and occipital regions in schizophrenia, including reduced regional homogeneity and disrupted temporal–occipital interactions \cite{jing2023deviant}. In particular, dysfunction in temporal cortical regions has been linked to impairments in auditory and language processing associated with psychotic symptoms \cite{zaykova2025lateralized, kaur2020structural}, while occipital regions have been implicated in abnormalities of visual processing and large scale functional connectivity \cite{liu2019occipital}. These findings provide a broader context for interpreting the observed stability of Temporal and Occipital representations in this study. However, within the present framework, such patterns should be understood as reflecting sensitivity of learned representations to region-aligned perturbations, rather than direct identification of disease-specific biomarkers. This highlights the potential of region-aware pretraining as a tool for probing functionally relevant spatial structure, while maintaining a clear distinction between representational behavior and underlying neurobiological mechanisms.

In contrast, the ADHD-200 dataset represents a more heterogeneous setting, with substantial variability in anatomical alignment across subjects and acquisition sites. Under such conditions, region specific interpretations become less direct, as patch level embeddings aligned to anatomically defined regions of interest may not correspond consistently across individuals. Consequently, observed regional patterns should be interpreted in terms of representation robustness rather than precise anatomical specificity. Despite this variability, strong performance is consistently observed for specific region–architecture combinations, particularly under the MA configuration. Limbic and Cerebellar regions demonstrate competitive performance across masking strategies, suggesting that representations aligned with these regions capture stable and transferable signals even in the presence of anatomical heterogeneity. The cerebellum has been widely implicated in ADHD, with converging evidence from neuroimaging and neurocognitive studies indicating its role in attention, executive function, and broader cognitive regulation, supported by its extensive connectivity with cortical and subcortical networks \cite{cundari2023neurocognitive}. In particular, cerebro–cerebellar circuits are increasingly recognized as central to neurodevelopmental processes underlying attention and behavioral control, and disruptions in these networks have been associated with core ADHD symptoms \cite{isaac2025can}. Similarly, alterations in limbic system function have been linked to emotional regulation and motivational processes, which are frequently affected in ADHD. Meta-analytic neuroimaging evidence further suggests that ADHD involves structural and functional changes in limbic and frontal systems, reflecting distributed network level dysregulation rather than isolated regional abnormalities \cite{yu2023meta}. Within this context, the observed stability of Limbic and Cerebellar representations suggests that region-aware masking may preferentially preserve signals associated with distributed cognitive and affective networks. However, these findings should be interpreted as reflecting the model’s ability to extract robust representations under heterogeneous conditions, rather than direct evidence of region-specific biomarkers. Overall, these results indicate that region-aware pretraining remains effective even when anatomical correspondence is less consistent, highlighting its robustness to real-world variability and its potential to capture functionally relevant patterns across diverse neuroimaging datasets.

The behavior of masking strategies further supports this interpretation. The comparable performance of \emph{Any} and \emph{Pure} in ADHD-200, and of \emph{Any} and \emph{Majority} in COBRE, indicates that different levels of spatial constraint can be beneficial depending on dataset characteristics. In this context, \emph{Pure} masking emphasizes more localized, region-specific signals, while \emph{Any} and \emph{Majority} facilitate the integration of distributed interactions across regions. These observations are consistent with the understanding that resting-state brain activity reflects both localized processing and large-scale network dynamics. Overall, these findings suggest that region aware masking should be viewed as a flexible mechanism for controlling the balance between localized and distributed representations, rather than as a direct indicator of anatomical importance. This distinction is particularly important in heterogeneous, multi-site datasets, where variability in acquisition and alignment limits the reliability of strict anatomical interpretations. More broadly, this perspective highlights that anatomically informed self-supervision can serve as a principled tool for probing functionally relevant brain organization while maintaining a clear separation between representation-learning behavior and underlying neurobiological interpretation.

\subsection{Limitations and Future Work}
While the proposed framework provides a systematic analysis of region-aware masking strategies and hybrid architectural design, several limitations should be considered. First, the current study operates on patch-based embeddings aligned with anatomically defined regions of interest, which introduces an abstraction over voxel-level structure and may limit the ability to capture fine-grained spatial variability. While this enables efficient training and structured incorporation of anatomical priors, it does not fully capture fine grained voxel level variability within regions. Extending the framework to voxel level masking could provide deeper insights into localized functional organization and more precise spatial representations. However, such approaches introduce substantial computational overhead and increased optimization complexity, particularly for high resolution 4D fMRI data. In addition, the current formulation relies on a fixed anatomical parcellation scheme, which may not optimally reflect task specific or data driven functional organization. Future work could explore alternative parcellation strategies, including functionally derived or hypothesis driven region definitions, to better align representation learning with specific neuroscientific objectives. Balancing representational granularity, parcellation design, and computational efficiency remains an important direction for future research. Second, the present study focuses on identifying effective masking strategies and architectural configurations for representation learning, rather than exhaustively optimizing models for each downstream task. While the MA architecture consistently delivers strong performance, and the proposed hybrid design yields substantial improvements over state-space-based approaches, it is important to note that transformer-based models such as SwiFT also achieve competitive performance in ADHD-200. This highlights that attention-based architectures remain highly effective for modeling global interactions in fMRI data, and further optimization or integration of transformer-based components may provide additional gains. A more detailed exploration of task-specific model adaptation, including systematic benchmarking across architectures and fine-tuning strategies, could further improve downstream outcomes and provide deeper insight into how pretrained representations transfer across datasets and conditions. 

Third, although region-aware masking introduces controlled spatial constraints, the current formulation does not explicitly model subject-specific variability or dynamic functional organization. This limitation is particularly important in the context of ADHD neuroimaging, where widely used multi-site datasets such as ADHD-200 exhibit substantial heterogeneity in acquisition protocols, cohort composition, and underlying neural patterns. Prior studies have demonstrated that findings derived from pooled datasets may not be consistent across individual cohorts, reflecting limited reproducibility and small effect sizes in resting-state biomarkers \cite{wang2017inconsistency}. Moreover, machine learning–based diagnostic frameworks are often sensitive to dataset-specific characteristics, leading to overfitting and reduced generalizability across sites and populations \cite{lanka2020supervised}. Review studies further highlight that variability in preprocessing pipelines, feature extraction strategies, and model selection contributes to inconsistent performance across ADHD detection studies \cite{ashraf2024systematic,taspinar2024review}. Within this context, the proposed framework partially addresses these challenges by systematically exploring multiple architectural configurations and region-aware masking strategies that impose varying degrees of spatial constraints and contextual mixing. By learning representations under diverse inductive biases, the model can capture complementary aspects of brain organization, potentially improving robustness to data heterogeneity and site-specific variability. Nevertheless, future work could extend this approach by incorporating adaptive or data-driven masking mechanisms that explicitly account for inter-subject variability and dynamic functional organization, thereby enhancing generalization in heterogeneous, multi-site neuroimaging settings. One promising direction is the development of a hierarchical, multi-scale architecture that integrates representations across different levels of brain organization. In such a framework, lower layers could operate at the voxel level to capture fine-grained spatial details, intermediate layers could model region-level structure through anatomically informed representations, and higher layers could encode network-level interactions reflecting large-scale functional connectivity. This multi-level design would enable the model to simultaneously capture local specificity and global context, potentially leading to more robust and biologically meaningful representations of fMRI data. Future work should also evaluate the framework on larger and more diverse datasets spanning other mental illnesses and neurological conditions, such as the SRPBS Multi-Disorder MRI Dataset, the REST-meta-MDD project for depression, the Transdiagnostic Connectome Project, and the Adolescent Brain Cognitive Development (ABCD) study.

The additional ablations also highlight the distinction between region-aware representation learning and dataset-level generalization. In COBRE, region-aware masking outperformed a matched random baseline for Any and Majority configurations, indicating that the benefit of anatomical masking cannot be explained by mask budget alone. In contrast, the ADHD-200 held-out-site experiment revealed a strong site-generalization bottleneck: all masking strategies and baseline models remained near or below chance-level AUROC. This suggests that while region-aware masking can improve the structure of learned representations, it is not sufficient by itself to overcome severe acquisition-site shift. Future work should therefore combine anatomically informed masking with explicit site-harmonization, domain-adaptation, or subject-specific alignment strategies.

More broadly, the proposed framework opens several avenues for extending self-supervised learning in neuroimaging. The ability to pretrain models with varying levels of spatial constraint, from heterogeneous to homogeneous masking, provides a flexible mechanism for tailoring representations to downstream tasks. This approach can be further extended to task-based fMRI, where variability across subjects and cognitive states introduces additional complexity and opportunities for learning more generalized representations. Finally, future research may explore more refined representation learning strategies guided by neuroscientific hypotheses, enabling closer alignment between computational models and brain organization. By combining anatomically informed masking with scalable pretraining and task-specific adaptation, this line of work has the potential to enable more robust, transferable modeling of large-scale neuroimaging data.
\section{Conclusion}
In this work, we introduced \textbf{Rhamba}, a region-aware self-supervised pretraining framework that integrates anatomically guided masking strategies with hybrid Attention–Mamba architectures for resting state fMRI analysis. Through systematic evaluation on downstream classification tasks using the COBRE and ADHD-200 datasets, we demonstrate that both masking design and architectural configuration play a central role in shaping learned representations. Our results reveal a clear distinction between reconstruction behavior and downstream performance. While stricter masking strategies simplify reconstruction, they do not necessarily improve classification representation quality. In parallel, hybrid architectures, particularly the MA configuration, achieve the strongest performance across datasets, highlighting the advantage of combining global attention with efficient state-space modeling to capture both distributed and long-range dependencies in fMRI data. Region-wise analysis further shows that masking primarily influences optimization dynamics rather than enforcing fixed anatomical importance. This suggests that effective representation learning benefits from a balance between localized structure and cross-regional interaction, rather than strictly localized constraints. Taken together, these findings emphasize that pretraining objectives and architectural design should be considered jointly when developing neuroimaging foundation models. More broadly, Rhamba provides a flexible framework for anatomically informed self-supervised learning, enabling systematic control over spatial constraints to adapt to different data characteristics and downstream tasks. The ADHD-200 held-out-site experiment further revealed site-generalization bottleneck, underscoring the challenge of cross-site transfer in heterogeneous multi-site rs-fMRI datasets. Future work may explore finer-grained or functionally derived parcellations, adaptive masking strategies, and extensions to task-based and multimodal neuroimaging settings. Such directions offer promising avenues toward more generalizable, interpretable, and scalable models for large-scale brain data analysis.

\section*{Conflict of Interest Statement}
The authors declare that they have no known competing financial interests or personal
relationships that could appear to influence the work reported in this paper.
 
\section*{Ethics Statement}

This study used only publicly available datasets, cited in the manuscript for full reference. As no new data were collected, no additional ethical approval was required. 
\section*{Author Contributions}
R.R.D. and P.P conceptualized the study, designed and implemented the analysis pipeline,
contributed to methodology development, and drafted the manuscript. P.E, C.T.R, M.J.S.
and R.S. provided critical revisions and intellectual input. R.S. supervised the project and
contributed to its conceptualization. All authors reviewed and approved the final version of
the manuscript. 

\section*{Funding}
This work was supported by the American Lebanese Syrian Associated Charities (ALSAC)
through salary support for P.P., C.T.R., and R.S., and internship stipends for R.R.D. and P.E.

\section*{Declaration of generative AI and AI-assisted technologies in the manuscript preparation process}
During the preparation of this work, the authors used ChatGPT 5.3 in order to improve the language and readability of the manuscript text. After using this tool, the authors reviewed and edited the content as needed and take full responsibility for the content of the published article.

\bibliographystyle{unsrtnat}   
\bibliography{references}      

\end{document}